\documentclass[twoside]{article}

\usepackage[preprint]{aistats2026}

\usepackage[round]{natbib}

\usepackage{microtype}
\usepackage{graphicx}
\usepackage{float}
\usepackage{wrapfig}
\usepackage{caption}
\usepackage{subcaption}
\usepackage{booktabs} %
\usepackage{bm}

\usepackage{algorithm2e}

\usepackage{amsmath}
\usepackage{amssymb}
\usepackage{mathtools}
\usepackage{amsthm}
\usepackage{thmtools}

\usepackage[disable]{todonotes}

\usepackage[pdfencoding=auto]{hyperref}

\usepackage[capitalize,noabbrev]{cleveref}

\DeclareMathOperator*{\argmin}{arg\,min}

\newcommand{\RR}{\mathbb{R}}

\newcommand{\X}{\mathcal{X}}
\newcommand{\Y}{Y}
\newcommand{\V}{\mathcal{V}}

\newcommand{\MMS}{\mathrm{MSIP}}
\newcommand{\MSIP}{\mathrm{MSIP}}

\newcommand{\F}{\mathcal{F}}

\newcommand{\Ffv}{{\mathcal{F}_{\delta}}}

\newcommand{\w}{\bm{w}}
\newcommand{\wi}{w_{i}}

\newcommand{\wit}{w_{i}^{(t)}}
\newcommand{\hw}{\hat{\bm{w}}}
\newcommand{\hwi}{\hat{w}_{i}}

\newcommand{\dotwit}{\dot{w}_{i}^{(t)}}
\newcommand{\witp}{w_{i}^{(t+1)}}

\newcommand{\wone}{w_{1}}

\newcommand{\hwone}{\hat{w}_{1}}

\newcommand{\wm}{w_{M}}

\newcommand{\hwm}{\hat{w}_{M}}

\newcommand{\wms}{w_{m}}
\newcommand{\wmst}{w_{m}^{(t)}}

\newcommand{\y}{\Y}

\newcommand{\yi}{y_{i}}
\newcommand{\yij}{y_{ij}}
\newcommand{\ymsj}{y_{mj}}
\newcommand{\yit}{y_{i}^{(t)}}
\newcommand{\yitp}{y_{i}^{(t+1)}}
\newcommand{\dotyit}{\dot{y}_{i}^{(t)}}

\newcommand{\dotz}{\dot{z}}

\newcommand{\yj}{y_{j}}
\newcommand{\yone}{y_{1}}

\newcommand{\ym}{y_{M}}
\newcommand{\yjt}{y_{m}^{(t)}}

\newcommand{\yt}{\Y^{(t)}}
\newcommand{\yms}{y_{m}}
\newcommand{\ymst}{y_{m}^{(t)}}

\newcommand{\dmut}{\dot{\mu}_t}
\newcommand{\mut}{\mu_{t}}

\newcommand{\ytplus}{\Y^{(t+1)}}

\newcommand{\xone}{x_{1}}
\newcommand{\xn}{x_{N}}

\newcommand{\xell}{x_{\ell}}
\newcommand{\xj}{x_{j}}

\newcommand{\vzero}{v_{0}}
\newcommand{\vzerov}{\bm{v}_{0}}

\newcommand{\bvzero}{\bar{v}_{0}}

\newcommand{\vonev}{v_{1}}
\newcommand{\vonem}{\bm{v}_{1}}

\newcommand{\hvonem}{\hat{\bm{v}}_{1}}

\newcommand{\bvonev}{\bar{v}_{1}}
\newcommand{\bvonem}{\bar{\bm{v}}_{1}}

\newcommand{\bmP}{\bm{P}(\y)}
\newcommand{\bmPt}{\bm{P}(\yt)}

\DeclareMathOperator{\MMD}{\mathrm{MMD}}

\newcommand{\Ky}{\bm{K}(\Y)}

\newcommand{\bmW}{\bm{W}(\Y)}
\newcommand{\bmWt}{\bm{W}(\yt)}
\newcommand{\what}{\hat{\bm{w}}(\Y)}
\newcommand{\whatt}{\hat{\bm{w}}(\yt)}
\newcommand{\whati}{\hat{w}_{i}(\Y)}
\newcommand{\whatms}{\hat{w}_{m}(\Y)}
\newcommand{\whatmst}{\hat{w}_{m}(\yt)}
\newcommand{\Dij}{\nabla_{ij}}
\newcommand{\dpix}{\,\mathrm{d}\pi(x)}
\newcommand{\bij}{\bm{b}_{ij}}
\newcommand{\Ki}{\bm{\bar{K}}_i}
\newcommand{\Kmbar}{\bm{\bar{K}}}

\newcommand{\DGS}[1]{\todo[color=green!40]{D\#: #1}}

\theoremstyle{plain}
\newtheorem{theorem}{Theorem}[section]
\newtheorem{proposition}[theorem]{Proposition}
\newtheorem{lemma}[theorem]{Lemma}
\newtheorem{corollary}[theorem]{Corollary}
\theoremstyle{definition}

\newtheorem{assumption}[theorem]{Assumption}
\theoremstyle{remark}

\crefname{corollary}{Corollary}{Corollaries}

\begin{document}

\runningtitle{From Mean Field to Mean Shift via Gradient Flows}

\runningauthor{Belhadji, Sharp, Marzouk}

\twocolumn[

\aistatstitle{Weighted Quantization Using MMD:\\ From Mean Field to Mean Shift via Gradient Flows}

\aistatsauthor{ Ayoub Belhadji \And Daniel Sharp \And Youssef Marzouk }
\aistatsaddress{ Center for Computational Science and Engineering\\ Massachusetts Institute of Technology } ]

\begin{abstract}
  Approximating a probability distribution using a set of particles is a fundamental problem in machine learning and statistics, with applications including clustering and quantization. Formally, we seek a weighted mixture of Dirac measures that best approximates the target distribution. While much existing work relies on the Wasserstein distance to quantify approximation errors, maximum mean discrepancy (MMD) has received comparatively less attention, especially when allowing for variable particle weights. We argue that a \textit{Wasserstein--Fisher--Rao} gradient flow is well-suited for designing quantizations optimal under MMD. We show that a system of interacting particles satisfying a set of ODEs discretizes this flow. We further derive a new fixed-point algorithm called \emph{mean shift interacting particles} (MSIP). We show that MSIP extends the classical mean shift algorithm, widely used for identifying modes in kernel density estimators. Moreover, we show that MSIP can be interpreted as preconditioned gradient descent and that it acts as a relaxation of Lloyd's algorithm for clustering. Our unification of gradient flows, mean shift, and MMD-optimal quantization yields algorithms %
more robust than state-of-the-art methods, as demonstrated via high-dimensional and multi-modal numerical experiments.

\end{abstract}

\section{INTRODUCTION}\label{sec:intro}

Numerous problems in statistics and machine learning involve approximating a probability measure using a small set of points, a task usually referred to as quantization \citep{GrLu00}. Examples include clustering \citep{Llo82}, numerical integration \citep{RoCaCa99,Bac17}, and teacher-student training of neural networks \citep{ChOyBa19,ArKoSaGr19}. Formally, the quantization problem consists of seeking an approximation to a probability measure $\pi$ using a mixture of Dirac measures $\sum_{i=1}^M \wi \delta_{\yi}$, where the $\yi$ %
are usually called nodes or centroids and the $\wi$ are weights. %
In this work, we focus on the problem 
\begin{equation}\label{eq:constrained_MMD_minimization}
\min_{\mu \in \mathrm{Mix}_{M}(\mathcal{X}) }\MMD\left(\pi,\mu \right),
\end{equation}
where $\mathrm{Mix}_{M}(\mathcal{X}):= \Big\{ \sum_{i=1}^{M} w_{i}\delta_{y_i}\,\big|\,  w_i \in \mathbb{R},\,y_i \in \mathcal{X} \Big\}$ and MMD is the maximum mean discrepancy defined as follows: given a measurable space $\mathcal{X}$ and a reproducing kernel Hilbert space (RKHS) $\mathcal{H}$, the MMD between measures $\mu$ and $\nu$ is defined as 
\begin{equation*}
\mathrm{MMD}(\mu,\nu) := \sup_{\|f\|_{\mathcal{H}}\leq 1}\Big|\int_{\mathcal{X}} f(x) \mathrm{d} \mu(x) - \int_{\mathcal{X}} f(x) \mathrm{d} \nu(x)\Big|.
\end{equation*}

 Initially introduced for statistical analysis and hypothesis testing \citep{GrFuTeScSm07, GrBoRaScSm12}, the MMD is applied in diverse fields including generative modeling \citep{DzRoGh15, LiChChYaPo17} and optimal transport \citep{ChNiRi24,PeCu19}.

\begin{figure*}[t]
    \centering
    \includegraphics[width=0.85\textwidth]{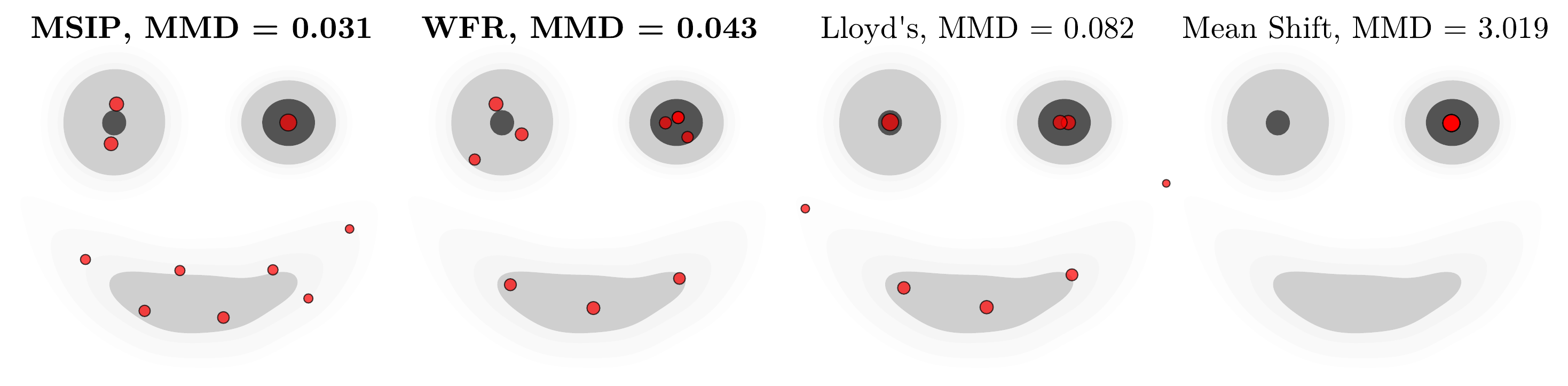}
    \caption{Comparison of quantization algorithms on a joker distribution, $M=10$. All algorithms initialized identically on the top-right mode. Each marker's size denotes the relative particle weight.}
    \label{fig:intro}
\end{figure*}

Lloyd's algorithm~\citep{Llo82} is a very popular quantization method, but seeks to minimize a Wasserstein distance instead of solving  problem~\eqref{eq:constrained_MMD_minimization}. %
Moreover, while many classical formulations of quantization constrain the weights ${w_i}$ to lie on the simplex, the minimization problem 
\eqref{eq:constrained_MMD_minimization} relaxes this requirement. %
Our approach leverages the natural connection between this minimization problem and gradient flows, a topic that has seen substantial theoretical and algorithmic progress in recent years. 
While MMD minimization using gradient flows has already been studied \citep{ArKoSaGr19,GlDvMiZh24}, most existing work considers the number of atoms tending to infinity, i.e., the mean-field limit. This conflicts with the fundamental motivation of quantization, and results in poor performance when the number of atoms is small. 
Our inspiration comes from considering problem \eqref{eq:constrained_MMD_minimization} 
at the opposite extreme, i.e., a single atom ($M = 1$). This corresponds to seeking the mode of a distribution using the \textit{mean shift} algorithm~\citep{FuHo75}. %
Mean shift is defined exclusively for this single particle case. Remarkably, %
the mean shift algorithm is a fixed-point iteration that we can also interpret as a preconditioned gradient ascent of the kernel density estimator (KDE), where the preconditioner helps the particle escape from low density regions. %

Building on these observations, we \textit{unify} views of quantization, gradient flows, and mode-seeking for MMD minimization using a \textit{finite} set of weighted particles. 
Concretely, our \textbf{main contributions} are as follows:

\begin{itemize}
    \item We tackle the minimization problem  \eqref{eq:constrained_MMD_minimization} by first adopting the perspective of gradient flows, where the MMD is a functional to be minimized by following the local direction of steepest descent in some chosen geometry. In particular, we propose using the Wasserstein--Fisher--Rao (WFR) geometry to allow for both transport and the creation and destruction of mass.

    \item We show that this approach admits efficient numerical implementation via an interacting particle system (IPS), described by ODEs for the time evolution of particle positions and weights. We characterize the infinite-time limit 
    of this WFR-IPS and introduce an algebraic system, depending on kernelized moments of the target $\pi$, that yields stationary solutions. %

    \item To directly find particles that satisfy this stationarity condition, we introduce a damped fixed-point algorithm termed \textit{mean shift interacting particles} (MSIP), which we show to extend the classical mean shift algorithm. We further show that MSIP is a \textit{preconditioned gradient descent} of 
    \[(y_1, \dots, y_M) \mapsto \min_{w_i \in \mathbb{R}} \frac{1}{2}\MMD^2 \left(\pi, \sum\limits_{i =1}^{M} w_{i} \delta_{y_i} \right),\]
    paralleling literature on Lloyd's algorithm \citep{DuFaGu99}. In other words, MSIP minimizes MMD analogously to how Lloyd's algorithm minimizes Wasserstein distance.%
    \item We demonstrate the computational \textit{efficiency} and \textit{robustness} of the proposed approaches for high-dimensional quantization problems, improving on current state-of-the-art methods. %
    In particular, both MSIP and the WFR-IPS yield \textit{near-optimal} quantizations under MMD, even with extremely adversarial initializations.
\end{itemize}

As a preview of our approach, \Cref{fig:intro} compares the quantizations obtained using four algorithms: MSIP, WFR-IPS, Lloyd's algorithm, and classical mean shift. We see that MSIP and WFR-IPS identify high-density regions of the distribution while capturing the anisotropy in all three modes. Mean shift, however, completely collapses to a single mode. While Lloyd's algorithm also captures anisotropy, it tends to capture the support of the \textit{entire} distribution instead of just high-probability regions. Finally, MSIP recognizes that the top-right mode is significantly concentrated and is adequately characterized with a single point; WFR-IPS and Lloyd instead over-represent variance within this mode. The quantizations' MMD values, shown at top, reflect these characteristics.

This article is structured as follows. In \Cref{sec:related_work}, we review the literature on numerical integration in an RKHS, quantization, the mean shift algorithm, and gradient flows. In \Cref{sec:algorithms} we present the main contributions of this work. In \Cref{sec:numerics}, we report numerical experiments that illustrate the proposed algorithms and compare to other schemes. \Cref{sec:discussion} is a short concluding discussion. %

\paragraph{Notation} We denote by $\mathcal{X}$ a subset of $\mathbb{R}^{d}$, and we use  $\mathcal{M}(\mathcal{X})$ and $\mathcal{P}(\mathcal{X})$ to refer to the set of measures and probability measures supported on $\mathcal{X}$, respectively. Moreover, given $M \in \mathbb{N}^{*}$, we denote by
$\mathcal{X}^{M}$ the $M$-fold Cartesian product of $\mathcal{X}$.%

We consider $\Y \in \mathcal{X}^{M}$ as a configuration of $M$ points $\yone, \dots, \ym \in \mathcal{X}$. Since $\mathcal{X} \subset \mathbb{R}^{d}$, we abuse notation and let $\Y$ refer to the $M \times d$ matrix whose rows are the vectors $\yone, \dots, \ym$. We also denote by $\Delta_{M-1}$ the simplex of dimension $M-1$ defined by
$\{ \bm{w} \in [0,1]^{M}| \:\: \sum_{i \in [M]} \wi = 1 \}$. Consider $\kappa:\mathcal{X}\times\mathcal{X}\to\mathbb{R}$ as the kernel associated to the RKHS $\mathcal{H}$. For a configuration $\Y$, we denote by $\bm{K}(\Y):= (\kappa(y_i,y_j))_{i,j \in [M]}$ the associated kernel matrix for $i,j \in [M]$. Additionally, for a given real-valued function $f: \mathcal{X} \rightarrow \RR$, we denote by $\bm{f}(\Y) := (f(y_i))_{i \in [M]}$ the vector of evaluations of $f$ on the elements of $\Y$. Finally, for a given vector-valued function $\bm{g}: \mathcal{X} \rightarrow \mathbb{R}^{d}$, we denote by $\bm{g}(\Y)$ the $M \times d$ matrix where each row is an evaluation of $\bm{g}$ on each $\yi$.

\section{BACKGROUND AND RELATED WORK}\label{sec:related_work}
The quantization problem \eqref{eq:constrained_MMD_minimization} is crucial in the design of efficient numerical integration methods. Indeed, for a given probability measure $\pi \in \mathcal{P}(\mathcal{X})$ and a function $f: \mathcal{X} \to \mathbb{R}$, a quadrature rule seeks to approximate $\mathbb{E}_\pi[f] := \int_{\mathcal{X}} f(x)\dpix$ with $\sum_{i = 1}^{M} \wi f(\yi)$, where the points $\yi \in \mathcal{X}$ are the quadrature nodes and the scalars $\wi$ are the quadrature weights. When the function $f$ belongs to the RKHS $\mathcal{H}$, the integration error is bounded as follows \citep{MuFuSrSc17}: 
\begin{equation}\label{eq:quadrature_mmd_inequality}
 \left|\mathbb{E}_\pi[f]%
 - \sum_{i = 1}^{M} \wi f(\yi) \right| \leq   \|f\|_{\mathcal{H}} \mathrm{MMD}\Big(\pi, \sum\limits_{i=1}^M \wi \delta_{\yi}\Big).
\end{equation}

In other words, having an accurate quantization with respect to MMD leads to a highly accurate quadrature rule $\sum_{i =1}^{M} \wi \delta_{\yi}$ for functions living in the RKHS. Within this framework, the design of the quadrature rule remains independent of the function $f$, which is particularly advantageous in scenarios where evaluating $f$ is expensive.

The \textit{optimal kernel-based quadrature} is a popular and well-studied family of quadrature rules: for a given configuration of nodes $\Y$, the weights $\hwone(\Y),\ldots,\hwm(\Y)$ satisfy 
\begin{equation}\label{eq:MMD_optimality_okq}
     \mathrm{MMD} \left( \pi, \sum\limits_{i=1}^M \hwi(\Y) \delta_{\yi}  \right) \leq \mathrm{MMD} \left( \pi, \sum\limits_{i=1}^M \wi \delta_{\yi} \right),
\end{equation}
for any $\wone, \dots, \wm \in \mathbb{R}$. These optimal weights can be expressed as the entries of the vector 
\begin{equation}\label{eq:optimal_w}
    \hat{\bm{w}}(\Y) = \bm{K}(\Y)^{-1} \vzerov(\Y),
\end{equation}
where $\vzero: \mathcal{X} \rightarrow \mathbb{R}$ is the \emph{kernel mean embedding} of the probability measure $\pi$ defined by 
\begin{equation}\label{eq:mke_def}
    \vzero(y):= \int_{\mathcal{X}} \kappa(x,y) \dpix.
\end{equation}
While the optimal weights $\hat{\bm{w}}(\Y)$ have an analytical formula for fixed configuration $\Y$, the function $\Y \mapsto \MMD^2\left(\pi, \sum_{i =1}^M \hwi(\Y) \delta_{\yi} \right)$
is daunting to minimize due to its non-convexity \citep{Oet17}. %
Recent work on optimal configurations often propose approaches tailored to specific standard probability measures %
\citep{KaSa18,KaSa19,EhGrOa19}. %
Some other methods are designed to be universal, but their numerical implementation is generally limited by difficulties in high-dimensional domains. Examples include ridge leverage score sampling \citep{Bac17}, determinantal point processes and volume sampling \citep{BeBaCh19, BeBaCh20,Bel21}, Fekete points \citep{KaSaTa21}, randomly pivoted Cholesky \citep{EpMo23}, or greedy selection algorithms \citep{De03,DeScWe05,SaHa16,Oet17,HuDu12,BrOaGiOs15,LaLiBa15,TeGoRiOa21}.

The function \eqref{eq:mke_def} is only practically tractable for some kernels and probability measures. One such case is when $\pi$ is an empirical measure, where the quantization can be seen as a summarization of $\pi$; this is %
notably relevant for Bayesian inference \citep{Owe17,RiChCoSwNiMaOa22}. %
There are many approaches to this problem. Kernel thinning~\citep{DwMa21,DwMa24} is mainly concerned with uniformly weighted quantization, where one can only choose $M$ if it satisfies $\lfloor N/2^r \rfloor$ for some positive integer $r$. Other methods such as kernel recombination \citep{HaObLy22, HaObLy23} or approximate ridge leverage score sampling \citep{ChScDeRo23} primarily focus on convergence rates in $M$, but are often outperformed in practice by simpler (e.g., greedy) methods when $M$ is small.

\subsection{Clustering}
A standard method for quantization \citep{Llo82} is Lloyd's algorithm, which aims to solve the problem
\begin{equation}\label{eq:w_2_quantization_problem}
    \min\limits_{\Y \in \mathcal{X}^{M}} \min\limits_{\bm{w} \in \Delta_{M-1}} \frac{1}{2}W_{2}^{2}\Big(\pi, \sum\limits_{i = 1}^{M} \wi \delta_{\yi} \Big),
\end{equation}
where $W_{2}$ is the $2$-Wasserstein distance \citep{PeCu19}.
The solution of \eqref{eq:w_2_quantization_problem} is naturally expressed in terms of \emph{Voronoi tessellations} \citep{DuFaGu99}. The Voronoi tesselation of a configuration $\Y \in \mathcal{X}^{M}$ corresponds to the sets $\mathcal{V}_{1}(\Y),\, \dots \,, \,\mathcal{V}_{M}(\Y)$ defined by $\mathcal{V}_{i}(Y) := \left\{\,x\in\mathcal{X}\ |\ i = \arg\min_{m\in[M]} \|x - y_m\|\,\right\}.$

The algorithm then consists of taking the fixed-point iteration $\ytplus = \bm{\Psi}_{\mathrm{Lloyd}}(\yt)$, where $\bm{\Psi}_{\mathrm{Lloyd}}: \mathcal{X}^{M} \rightarrow \mathcal{X}^{M}$ is the map defined by 
\begin{equation}\label{eq:lloyd_map}
\forall i \in [M], \:\:  \left(\bm{\Psi}_{\mathrm{Lloyd}}(\Y)\right)_{i, :} := \frac{\int_{\V_{i}(\Y)}x \mathrm{d} \pi(x)}{\int_{\V_{i}(\Y)}\mathrm{d} \pi(x)} .
\end{equation}

The fixed points of $\bm{\Psi}_{\mathrm{Lloyd}}$ are critical points of the function $G_M$, defined as \citep[Proposition 6.2.]{DuFaGu99}
\begin{equation}
  G_{M}: \Y \mapsto \min\limits_{\bm{w} \in \Delta_{M-1}} \frac{1}{2}W_{2}^{2}\Big(\pi, \sum\limits_{i \in [M]} \wi \delta_{\yi} \Big).
\end{equation}
The authors prove that $G_{M}$ is differentiable in $\Y$ with the gradient given by
\begin{equation}\label{eq:gradient_G_M_lloyd}
    \nabla G_{M}(\Y) = \bm{D}(\Y) (\Y - \bm{\Psi}_{\mathrm{Lloyd}}(\Y)),
\end{equation}
where $\bm{D}(\Y) \in \mathbb{R}^{M \times M}$ is the diagonal matrix $(\bm{D}(\Y))_{i,i} := \int_{\V_{i}(\Y)} \dpix$. Hence, the fixed-point iteration is equivalent to preconditioned gradient descent, i.e., 
\begin{equation}
    \ytplus = \yt - \bm{D}(\yt)^{-1} \nabla G_{M}(\yt).
\end{equation}

The theoretical analysis and nondegeneracy of Lloyd's algorithm has been the subject of extensive study \citep{Kie82,Wu92,DuFaGu99,DuEmJu06,EmJuRa08,PaYu16,PoCaPa24}.

\subsection{Mean shift}\label{sec:mean_shift}
Mean shift is a widely used algorithm for tasks such as clustering \citep{FuHo75,Car15} and image segmentation \citep{CoMe02}. Mean shift locates the mode of a kernel density estimator (KDE) %
\begin{equation}
\hat{f}_{\mathrm{KDE}}(y):= \frac{1}{N} \sum\limits_{\ell=1}^{N}\kappa(y,\xell) = \vzero(y),
\end{equation}
where $\vzero$ is defined by \eqref{eq:mke_def}, taking $\pi$ to be the empirical measure associated to the data points $\xone, \dots, \xn  \in \mathcal{X}$. In this context, the kernel $\kappa$ must satisfy the following. 
\begin{assumption}\label{assumption:gradient_kappa} The kernel $\kappa$ is $\mathcal{C}^1$ and there is a symmetric kernel $\bar{\kappa}: \mathcal{X} \times \mathcal{X} \rightarrow \mathbb{R}$ satisfying
\begin{equation}\label{eq:gradient_condition_msip}
\forall x,y \in \mathcal{X}, \quad \nabla_{2}\kappa(x,y) = (x-y) \bar{\kappa}(x,y).
\end{equation}
\end{assumption}
Under \cref{assumption:gradient_kappa}, a critical point $y_{\star}$ of the KDE satisfies $\sum_{\ell=1}^N (y_{\star}-\xell)\bar{\kappa}(\xell,y_{\star}) = 0 $, implying
\begin{equation}\label{eq:ms_equation}
    y_{\star} = \bm{\Psi}_{\mathrm{MS}}(y_{\star}):= \frac{\sum_{\ell=1}^N \xell \, \bar{\kappa}(\xell,y_{\star})}{\sum_{\ell=1}^N \bar{\kappa}(\xell,y_{\star})}.
\end{equation}
The mean shift algorithm seeks the location of $y_{\star}$ using a fixed-point iteration on map $\bm{\Psi}_{\mathrm{MS}}$. Convergence of mean shift is extensively investigated in \citep{Che95,LiHuWu07,Car07,Gha13,Gha15,ArMaPe16,YaTa19,YaTa24}. \Cref{assumption:gradient_kappa} is satisfied by \textbf{many distance-based kernels} defined as $\kappa(x,y) = \varphi(\|x-y\|)$ for some positive function $\varphi\in\mathcal{C}^1(\RR^+)$. This includes the squared exponential kernel, the inverse multiquadric kernel (IMQ), and differentiable Matérn kernels.

\subsection{Gradient flows for MMD minimization}

A gradient flow of a functional $\mathcal{F}$ on the space of positive measures $\mathcal{M}_{+}(\mathcal{X})$ is a trajectory $\mu_{t}$ satisfying the partial differential equation (PDE) 
\begin{equation}\label{eq:abstract_gradient_flow}
   \dmut = -\mathrm{grad}_D \mathcal{F}[\mut],
\end{equation}
where $\mathrm{grad}_D \mathcal{F}$ denotes the gradient of $ \mathcal{F}$ with respect to a metric $D: \mathcal{M}_{+}(\mathcal{X}) \times \mathcal{M}_{+}(\mathcal{X}) \rightarrow \mathbb{R}$;
see \citep{Ott01,AmGiSa08,San17,ChNiRi24} for details. %
Gradient flows are extensively used in machine learning and statistics, particularly for sampling \citep{JoKiOt98,CrBe16,CaCrPa19,ChGoLuMaRi20,SaKoLu20,GlArGr21,KoAuMaAb21,LuSlWa23,MaMa24}.

For MMD minimization, the functional $\F$ and its first variation $\Ffv$ are taken to be 
\begin{subequations}\label{def:mmd}
\begin{equation}
    \F(\mu) := \frac{1}{2} \mathrm{MMD}(\mu, \pi)^2,
\end{equation}
\begin{equation}
    \Ffv[\mu](\cdot) = \int_{\X} \kappa(\cdot,x)\,\mathrm{d}\mu(x) - \int_{\X} \kappa(\cdot,x)\dpix.
\end{equation}
\end{subequations}
The functional $\Ffv$ quantifies how $\F$ changes due to a perturbation $\chi \in \mathcal{P}(\mathcal{X})$ around a measure $\mu$, %
and is an essential object in the gradient flows literature. \Cref{tab:gf_pdes} summarizes various choices of metric $D$ that have been used for MMD gradient flows, along with the corresponding PDE. We horizontally align corresponding terms

\begin{table*}[t]
\centering
\renewcommand{\arraystretch}{0.7}
\caption{Gradient Flow Geometries for Minimizing MMD Functional $\F$}%
\label{tab:gf_pdes}
\begin{tabular}{@{ }c@{\hskip.05cm}lrrr@{ }}
\toprule
\textbf{Geometry metric $D$} & \multicolumn{4}{c}{\textbf{PDE} $\dmut = -\mathrm{grad}_D \mathcal{F}[\mut]$} \\
\midrule
Wasserstein \citep{ArKoSaGr19,AlHeSt23} &
$\dmut =$ & $\alpha\ \mathrm{div} \left( \mut\nabla \Ffv [\mut] \right)$ \\[0.5em]
Fisher--Rao &$\dmut =$ & & $-\beta\,\Ffv[\mut] \mut$ \\[0.5em]
MMD & $\dmut =$ & & & $-\gamma(\mu_{t} - \pi)$ \\[0.5em]
IFT \citep{GlDvMiZh24,ZhMi24}&
$\dmut =$ & $\alpha\ \mathrm{div} \left( \mu_{t} \nabla \Ffv [\mut] \right)$ & &$ - \gamma(\mut - \pi)$ \\[0.5em]
WFR \citep{GlDvMiZh24} &
$\dmut =$ & $ \alpha\ \mathrm{div}\left(\mut \nabla \Ffv[\mut]\right) $& $- \beta\,\Ffv[\mut] \mut$ \\[0.5em]
\bottomrule
\end{tabular}
\end{table*}

Here, we specifically consider literature that discretizes the dynamics of $\mu_t$ 
in \eqref{eq:abstract_gradient_flow} using interacting particles, 
but also include the MMD\footnote{Not to be confused with when $\F$ is the MMD.} and Fisher--Rao geometries for completeness; with this mean-field ODE perspective, however, these latter two geometries have limited use. \citet{ArKoSaGr19} study the gradient flow of $\F$ under the 2-Wasserstein ($W_2$) geometry, %
while \citep{GlDvMiZh24} proposes inheriting properties from both the $W_2$ and MMD geometries via \emph{spherical interaction-force transport} (IFT).
Despite being able to use interacting particle systems, both $W_2$ and IFT flows of the MMD suffer from significantly degraded performance when initialized poorly. In particular, they require a large number of particles to converge, making the resulting quantization inefficient for, e.g., quadrature.

Finally, the Wasserstein-Fisher–Rao (WFR) geometry %
allows both mass transport and total mass variation, %
extending the $W_2$ distance to measures with different total masses, i.e., \emph{unbalanced} optimal transport \citep{KoMoVo16,GaMo17,ChPeScVi18,LiMiSa18}.\footnote{While we use the convention from~\citep{GaMo17},~\citet[Remark 2.2]{MiZh25} address the historical differentiation of the Fisher--Rao and closely-related Hellenger geometries.}
Compared to the Wasserstein gradient flow, the WFR gradient flow introduces a reaction term, which allows the total mass of $\mu_t$ to change as a function of $t$. %
Despite this formulation, the numerical scheme proposed by \citep{GlDvMiZh24} to solve this PDE artificially conserves mass by projecting the weights onto the simplex, inconsistent with the PDE's theoretical properties.%

\section{MAIN RESULTS}\label{sec:algorithms}
This section is structured as follows: %
In \Cref{sec:wfr}, we discretize a WFR gradient flow using a system of ODEs describing an interacting particle system (IPS), then show that the MMD decreases along the ODE solution. Moreover, we characterize conditions for the steady-state solution of these ODEs. In \Cref{sec:msip}, we derive \textit{mean shift interacting particles} (MSIP) as a fixed-point iteration for satisfying steady-state conditions of the WFR gradient flow. Then, we show that MSIP can be expressed as a preconditioned gradient descent on the MMD as a natural extension of the mean shift algorithm. \cref{alg:wfr_gf,alg:msip} in \cref{sec:alg_lists} describe implementations of WFR-IPS and MSIP.

\subsection{An IPS approach to WFR gradient flows}\label{sec:wfr}
We consider the WFR gradient flow for minimizing the MMD (cf.\ \cref{tab:gf_pdes}), setting the reaction speed to $\beta\equiv 1$ without loss of generality:
\begin{equation}\label{eq:WFR_for_MMD}
    \dmut =\alpha\ \mathrm{div}\left(\mut \nabla \Ffv[\mut]\right) - \,\Ffv[\mut] \mut.
\end{equation}
We describe $\mut$ via $M$ Diracs, i.e.,~\(\mut = \sum_{i = 1}^{M} \wit\delta_{\yit},\) %
for $\yit\in \mathcal{X} $ and $\wit \in \RR$. 

\begin{proposition}\label{prop:coupled_ode_WFR}
Define the system of ordinary differential equations
\begin{align}\label{eq:WFR_particle_equation}
    &\dotyit = - \alpha\nabla\Ffv[\mu_t](\yit),\ \dotwit = -  \wit\Ffv[\mut](\yit)\nonumber\\
    &\Ffv[\mut](y) = \sum\limits_{m = 1}^{M} \wmst \kappa(\yjt, y) - v_{0}(y) 
\end{align}
where $i \in [M]$, $\alpha > 0$, and $\vzero$ is defined in \eqref{eq:mke_def}.
If $(\mut)_{t \geq 0}$ solves \eqref{eq:WFR_particle_equation}, then it weakly satisfies \eqref{eq:WFR_for_MMD}.
\end{proposition}
We can then use standard ODE solvers to efficiently approximate $\mut$, provided the function $\vzero$ can be evaluated and differentiated. When $\pi$ is an empirical measure of $N$ atoms in $\RR^d$, this simply requires differentiability of kernel $\kappa$, and simulating \eqref{eq:WFR_particle_equation} is
$O(d(M+N)M)$ operations.%
\footnote{This assumes evaluating $\kappa$ is $O(1)$ and $\nabla_2\kappa$ is $O(d)$.} %
The proof and a more detailed account of \Cref{prop:coupled_ode_WFR} are given in \Cref{proof:coupled_ode_WFR}.
\begin{proposition}\label{prop:wfr_convergence}
Assume that $\kappa\in\mathcal{C}^{1}$ has bounded gradient,  $\kappa(x,y) \geq 0$ for any $x,y \in \mathcal{X}$, and  $y\mapsto\kappa(y,y)$ is a constant function with value $B_{\kappa}>0$. %
Let $y_i \in \mathcal{X}$ and $w_i \in (0,1)$ for $i\in[M]$.  Then there exists a unique solution of \eqref{eq:WFR_particle_equation} over $t\in[0,+\infty)$ with $y_{i}^{(0)} = y_i$ and $w_{i}^{(0)} = w_i$. Further, we have $w_{i}^{(t)} \in (0,1)$ for $t \geq 0$, and $t_2 \geq t_1 \geq 0$ implies
    \begin{equation}
        \mathcal{F}(\mu_{t_2}) \leq \mathcal{F}(\mu_{t_1}).
    \end{equation}

\end{proposition}%
The
proof of \Cref{prop:wfr_convergence} is given in \Cref{proof:wfr_convergence}. The system of ODEs \eqref{eq:WFR_particle_equation} yields quadrature rules with bounded non-negative weights, which is a desirable property in the context of numerical integration \citep{KaKaSa19}. 
As $\F$ is bounded from below, we know that \eqref{eq:WFR_particle_equation} approaches a stationary point as $t\to\infty$. %
We are then primarily interested in the minimization of the MMD when the particles and weights reach stationarity as well, i.e., $\dotyit=0$ and $\dotwit=0$ as $t\to\infty$, which is not necessarily guaranteed by \cref{prop:wfr_convergence}. %
Conditions on $\yi$ and $\wi$ sufficient for stationarity are:
\begin{equation}\label{eq:mixture_diracs_kbar_eq_1}
       \vzero(\yi) =\sum\limits_{m = 1}^{M} \wms \kappa(\yms,\yi), 
\end{equation}
\begin{equation}\label{eq:MMD_gradient_mixture}
       \nabla\vzero(\yi) = \sum\limits_{m = 1}^{M} \wms \nabla_{2} \kappa(\yms,\yi). %
\end{equation}
Condition~\eqref{eq:mixture_diracs_kbar_eq_1} is prevalent in the literature on kernel-based quadrature, and reflects that the quadrature rule defined by the measure $\mu = \sum_{m = 1 }^{M} \wms \delta_{\yms}$ is exact on the subspace spanned by the functions $\kappa(\cdot,\yi)$, equivalent to \eqref{eq:optimal_w}. %
Above, we have made few assumptions on the structure of $\kappa$. We now use \cref{assumption:gradient_kappa} to explore condition~\eqref{eq:MMD_gradient_mixture}, less known in the literature.

\begin{proposition}\label{prop:steady_state_equation_on_y}
Let $\y \in \mathcal{X}^{M}$. Under \Cref{assumption:gradient_kappa} and assuming that the gradient of $\kappa$ is bounded on $\X$,  \eqref{eq:mixture_diracs_kbar_eq_1} and \eqref{eq:MMD_gradient_mixture} imply that 
\begin{equation}\label{eq:mixture_diracs_kbar_eq_2_reformulation}
     \bm{\bar{K}}(\y)\bm{W}(\y)\y  = \hvonem(\Y) ,
\end{equation}
where $\bm{\bar{K}}(\y) \in \RR^{M \times M}$ is the kernel matrix with kernel $\bar{\kappa}$, the diagonal matrix $\bm{W}(\y)$ has entries corresponding to those of $\hw(\y)$ given by \eqref{eq:optimal_w}, and $\hvonem(\y) \in \RR^{M \times d }$ is the matrix with rows defined as
\begin{equation}
\label{eq:v_hat_def}
(\hvonem(\y))_{i,:} :=  \nabla v_0(\yi) + \yi \sum_{m=1}^{M} \hat{w}_m\bar{\kappa}(\yms,\yi).
\end{equation}
\end{proposition}
When using the squared-exponential kernel with unit bandwidth, \eqref{eq:mixture_diracs_kbar_eq_1}, \eqref{eq:MMD_gradient_mixture}, and \eqref{eq:v_hat_def} give
\begin{equation}\label{eq:vone_def}
    (\hvonem(\Y))_{i,:} = \vonev(\yi) := \int x\kappa(x,\yi)\dpix.
\end{equation}
Thus $\vonev$ resembles a ``kernelized first moment'' (hence the subscript), similar to how $\vzero(\yi)$, the KDE of $\pi$, resembles a ``kernelized zeroth moment.'' The proof of \Cref{prop:steady_state_equation_on_y} is given in \Cref{proof:steady_state_equation_on_y}.

\subsection{A fixed-point scheme for steady-state solutions}\label{sec:msip}

In general, we cannot analytically find $\y$ satisfying \eqref{eq:mixture_diracs_kbar_eq_2_reformulation}. Instead, consider the sequence $(\yt)$ with
\begin{equation}\label{eq:msip_fixed_point_def}
     \bm{\bar{K}}(\yt) \bm{W}(\yt) \ytplus = \hvonem(\yt).
\end{equation}
Calculating $\bm{W}(\yt)$ of course requires the invertibility of $\bm{K}(\yt)$. If $\bm{\bar{K}}(\yt)$ is also nonsingular, \eqref{eq:msip_fixed_point_def} is a fixed-point iteration, $\ytplus =  \bm{\Psi}_{\MSIP}(\yt)$, with
\begin{equation}\label{eq:fixed_point_general_case}
    \bm{\Psi}_{\MSIP}(\Y): = \bm{W}(\Y)^{-1} \bm{\bar{K}}(\Y)^{-1} \hvonem(\Y).
\end{equation}

We call $\bm{\Psi}_{\MSIP}$ the \textit{mean shift interacting particles} map.
We now show that iteration of $\Psi_{\MMS}$ can be seen as a preconditioned gradient descent of the function 
$F_M: \mathcal{X}^{M} \rightarrow \mathbb{R}$ given by
\begin{equation}\label{eq:opt_mmd_functional}
    F_{M}(\yone, \dots, \ym) := \inf\limits_{\w \in \mathbb{R}^{M}} \F \left[ \sum\limits_{i=1}^{M}\wi \delta_{\yi} \right].
\end{equation}

\begin{theorem}\label{thm:gradient_opt_mmd}
    Let $Y\in\X^M$ be such that $\det\Ky > 0$ and define $\bm{P}(\Y) :=\bm{W}(\y) \bar{\bm{K}}(\y) \bm{W}(\y)$. Under \Cref{assumption:gradient_kappa}, we have
\begin{align}\label{eq:gradien_opt_mmd_under_assumption}
\nabla F_{M}(\y) &=  \bm{W}(\y)\left(\bar{\bm{K}}(\y)\bm{W}(\y)\y - \hat{\bm{v}}_1(\y) \right)\nonumber\\
&= \bm{P}(\y) \Big( \y - \bm{\Psi}_{\MMS}(\y) \Big)
\end{align}
with $\nabla F_M:\X^M\to\RR^{M\times d}$ and $\bm{\Psi}_{\MMS}(\y)$ from \eqref{eq:fixed_point_general_case} when $\bm{P}(\y)$ is invertible.

\end{theorem}
\Cref{thm:gradient_opt_mmd} is proved in \cref{proof:thm_gradient_opt_mmd}. The identity \eqref{eq:gradien_opt_mmd_under_assumption} proves that the fixed points of the function $\bm{\Psi}_{\MMS}$ are critical points of the function $F_M$. This result is reminiscent of \eqref{eq:gradient_G_M_lloyd}, proved in \citep{DuFaGu99}, which gives a similar characterization of the fixed points of Lloyd's map $\bm{\Psi}_{\mathrm{Lloyd}}$. Following identical reasoning as \citep{PoCaPa24}, %
we deduce from~\eqref{eq:gradien_opt_mmd_under_assumption} that the fixed point iteration defined by \eqref{eq:fixed_point_general_case} is a preconditioned gradient descent, with $\bmP$ as a preconditioning matrix. If matrix $\bmPt$ is nonsingular, \eqref{eq:fixed_point_general_case} and \eqref{eq:gradien_opt_mmd_under_assumption} yield
\begin{equation}\label{eq:fixed_point_as_precond_grad_descent}
  \ytplus   = \yt - \bmPt^{-1} \nabla F_{M}(\yt).
\end{equation}
If we choose a step size $\eta$ for \eqref{eq:fixed_point_as_precond_grad_descent}, we obtain a damped fixed-point iteration~\citep{Ke18} on $\bm{\Psi}_{\MMS}$ as
\begin{align}\label{eq:fixed_point_as_precond_grad_descent_alpha}
  \ytplus &= \yt - \eta\bmPt^{-1} \nabla F_{M}(\yt) \nonumber\\
  &= (1-\eta) \yt + \eta \bm{\Psi}_{\MMS}(\yt).
\end{align}

In the case of $\kappa\propto\bar{\kappa}$, i.e., %
squared-exponential~\citep{Che95}, we can use $\vonem$ from \eqref{eq:vone_def} to simplify \eqref{eq:fixed_point_as_precond_grad_descent_alpha} into
\begin{equation}\label{eq:msip_for_gaussian}
    \ytplus = (1-\eta)Y^{(t)} + \eta\bm{W}(\yt)^{-1} \bm{K}(\yt)^{-1} \vonem(\yt).
\end{equation}
The following result shows the existence of a step size schedule for which the MMD decreases along the iterates of \eqref{eq:msip_for_gaussian}.
\begin{proposition}\label{prop:mmd_inimization_using_msip}
    Let $\kappa$ be a squared-exponential kernel. For any $Y^{(0)}$ such that $\hat{\bm{w}}(\y^{(0)})$ exists and has strictly positive entries, there exist positive step sizes $(\eta_{t})_{t\geq 0}$ such that the sequence
    \begin{equation}
        \ytplus = (1-\eta_t) \yt + \eta_t \bm{W}(\yt)^{-1} \bm{K}(\yt)^{-1} \vonem(\yt)
    \end{equation}
    satisfies $F_{M}(\ytplus) \leq F_{M}(\yt)$ for $t \geq 0$, with invertible $\bm{K}(\yt)$ and positive $\hat{\bm{w}}(\yt)$. 
\end{proposition}%
\begin{proposition}\label{prop:exist_Y}
    Under the assumptions of \cref{prop:mmd_inimization_using_msip} and for unweighted empirical $\pi$, if $\y^{(0)}$ are distinct points on the support of $\pi$, then there exists a bandwidth of $\kappa$ for which $\hat{\bm{w}}(\y^{(0)}) > 0$. 
\end{proposition}
The proofs of \Cref{prop:mmd_inimization_using_msip,prop:exist_Y} are given in \Cref{proof:mmd_inimization_using_msip,app:proof_exists_config}. %
\Cref{prop:mmd_inimization_using_msip} yields a descent property for \eqref{eq:fixed_point_as_precond_grad_descent_alpha}, allowing for varying step sizes. %
\Cref{prop:mmd_inimization_using_msip,prop:exist_Y} 
resemble early existence and convergence guarantees for Lloyd's algorithm, refined over many decades~\citep{DuEmJu06,PaYu16,PoCaPa24}. Further, the bandwidth consideration in \cref{prop:exist_Y} is common in kernel methods~\citep{Chi91}. While \cref{prop:mmd_inimization_using_msip} requires initially positive weights for theoretical purposes, we have observed empirical success when initializing a configuration with negative weights. Extending \cref{prop:mmd_inimization_using_msip,prop:exist_Y} to arbitrary initialization and fixed step size is left to future work: guaranteeing convergence for a fixed step size requires higher-order analysis of the function $F_M$ and is beyond the scope of this work.

\subsection{Mean shift as an accelerated gradient flow}
In the case of one particle, i.e., $M = 1$, we show that the classical mean shift algorithm identifies critical points of the function $F_1$ given in  \eqref{eq:opt_mmd_functional}.

\begin{corollary}\label{cor:ms_as_MSIP} When $\pi$ is an empirical distribution, under \Cref{assumption:gradient_kappa} and further assuming $\bar{v}_{0}(y):= \int_{\X} \bar{\kappa}(x,y) \dpix \neq 0$ and $v_0(y)\neq 0$, the following expressions are equal
\begin{align}\label{eq:gradient_F_1}
    \nabla F_{1}(y) &\equiv \frac{\vzero^2(y)\bar{\kappa}(y,y)}{\kappa^2(y,y)}(y-\bm{\Psi}_{\MMS}(y)) = \nonumber\\
    &\equiv \frac{v_{0}(y)\bar{v}_0(y)}{\kappa(y,y)}(y - \bm{\Psi}_{\mathrm{MS}}(y)),
\end{align}
where $\bm{\Psi}_{\mathrm{MS}}$ is defined by \eqref{eq:ms_equation}. In particular, when $\bar{\kappa} = \lambda\kappa$ for $\lambda \neq 0$,  $\bm{\Psi}_{\MMS}(y) = \bm{\Psi}_{\mathrm{MS}}(y).$
\end{corollary}
\Cref{cor:ms_as_MSIP} shows the intimate connections between the mean shift interacting particles and classical mean shift maps. 
In the simple case of squared-exponential $\kappa$ and a single particle, the maps are the same.
More broadly, for any critical point $y$ of $F_1$, satisfying the assumptions of \cref{cor:ms_as_MSIP} automatically yields that $y$ is a critical point of \textit{both} $\bm{\Psi}_{\MMS}$ and $\bm{\Psi}_{\mathrm{MS}}$. %
The proof of \Cref{cor:ms_as_MSIP} is given in \Cref{proof:mean_shift_as_mmd_min} as a direct consequence of \Cref{thm:gradient_opt_mmd}.

In this single particle setting, we contrast the definition of MSIP \eqref{eq:msip_for_gaussian} with an unweighted Wasserstein gradient flow for the MMD~\citep{ArKoSaGr19}. For simplicity, we consider squared-exponential $\kappa$ with unit bandwidth, unifying MSIP and mean shift. The algorithm derived in \citep{ArKoSaGr19} boils down to the iteration 
\begin{align}
    y^{(t+1)} &= y^{(t)} + \eta \nabla \vzero(y^{(t)})\nonumber\\
    &= \big(1-\eta  \vzero(y^{(t)}) \big)y^{(t)} + \eta\vonev(y^{(t)}).
\end{align}
When $y^{(0)}$ is far from the data, small $\vzero(y^{(t)})$ and $\|\vonev(y^{(t)})\|$ give practically still particles: $y^{(1)} \approx y^{(0)}$. The preconditioner in \eqref{eq:msip_for_gaussian}, however, prevents mean shift steps from vanishing in unlikely regions. Thus, we expect mean-shift to \textbf{accelerate steady-state behavior}\DGS{wording?} over typical gradient-flow methods. \Cref{sec:ms_vs_ga} and \Cref{fig:ms_vs_ga} give additional details on this phenomenon.

\section{NUMERICAL RESULTS}\label{sec:numerics}
\begin{figure}[t!]
    \centering
    \includegraphics[width=0.99\linewidth]{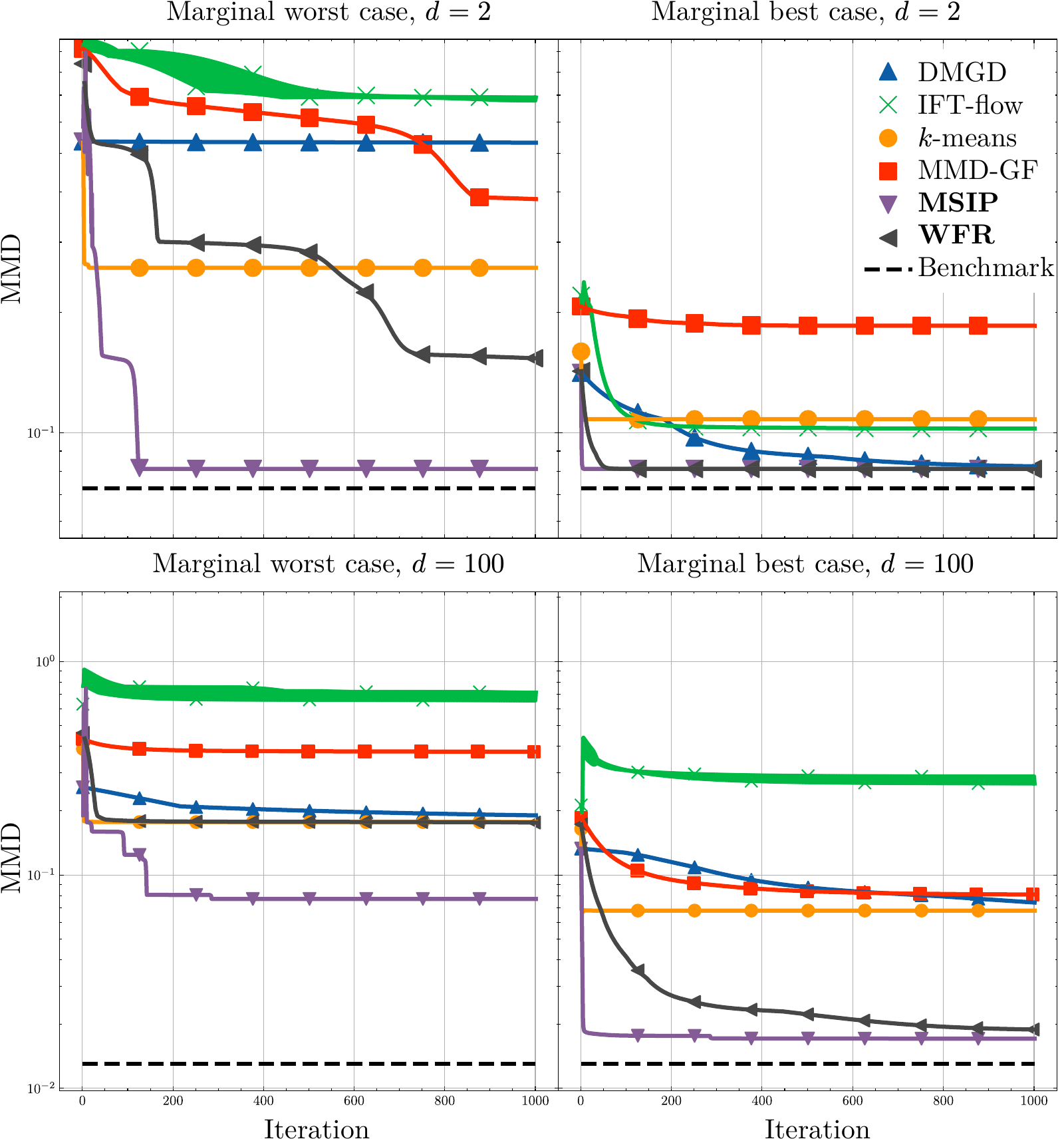}
    \caption{Comparison of different quantization algorithms on a GMM. (Top): dimension $d=2$, $L_0 = M = 3$. (Bottom): dimension $d=100$, $L_0=5$, $M=10$. We use squared-exponential kernel with $\sigma=5$ for kernelized algorithms with hyperparameter tuning.}%
    \label{fig:mmd_comparison_gmm}
\end{figure}

We now summarize numerical simulations that validate the proposed algorithms.%
\footnote{\href{https://github.com/AyoubBelhadji/disruptive_quantization}{github.com/AyoubBelhadji/disruptive\_quantization}}
In \Cref{sec:synthetic_basic_numerics} we conduct numerical experiments on Gaussian mixture datasets, while in \Cref{sec:real_datasets_basic_numerics}, we perform the experiments on MNIST.  %
We provide additional details about these numerics in \cref{app:implementation_details,sec:settings_numerics_GMM,appendix:viz_paths_synthetic_dataset}, comparisons to kernel thinning~\citep{DwMa24} in \cref{sec:app_thinning}, examples for many distributions in \cref{sec:app_examples}, and resulting weights of our algorithms in \cref{sec:app_weights}. We demonstrate further practicality for data science by clustering high-dimensional features of cellular images in \cref{app:dataset}.

\subsection{Experiment 1: Gaussian mixture datasets}\label{sec:synthetic_basic_numerics}
We consider quantization of two Gaussian mixture distributions, represented as empirical measures of $N=1000$ atoms: a mixture of $L_0 = 3$ Gaussians in $d=2$ and a mixture of $L_0 = 5$ Gaussians in $d=100$. We compare the following algorithms: i) Wasserstein--Fisher--Rao Flow (WFR) from \Cref{sec:wfr}, ii) mean shift interacting particles (MSIP) from \Cref{sec:msip}, iii) IFTflow \citep{GlDvMiZh24}, iv) MMD gradient flow (MMDGF) \citep{ArKoSaGr19}, v) $k$-means, vi) discrepancy-minimizing gradient descent (DMGD) on the function $F_{M}$ in \eqref{eq:opt_mmd_functional} without preconditioning. %
\Cref{cor:ms_as_MSIP} suggests MSIP will outperform DMGD.

\Cref{fig:mmd_comparison_gmm} shows MMD as a function of iteration number for quantizations obtained from running each algorithm on 100 random initializations. %
At each iteration and for each algorithm, we plot the highest and lowest MMD value across all 100 initializations. %
We observe that, even in the worst case, MSIP consistently outperforms other quantization algorithms, quickly converging to a near-optimal quantization under the MMD. %
The latter is captured by the dashed line, which is a standard benchmark for the MMD of optimal quadrature with $M$ points; see details in \Cref{app:implementation_details}. 
For $d=2$, WFR more slowly optimizes the MMD, eventually surpassing $k$-means. In the best case, all algorithms perform well except for MMDGF. When $d=100$, we have a more difficult approximation problem, with MSIP performing best in both the best- and worst-case instances at each iteration. WFR performs comparably to $k$-means in the worst case and to MSIP in the best case. Other algorithms struggle in this high-dimensional problem.

\textit{Qualitatively}, \cref{fig:all_trajectories} in \cref{appendix:viz_paths_synthetic_dataset} further illustrates the sensitivity of each algorithm to its initialization.
MMDGF, IFTFlow, and DMGD are not robust to initializations in the two-dimensional example, whereas MSIP and WFR perform well even when initialized far from the target support; $k$-means lies in between.

\subsection{Experiment 2: MNIST} \label{sec:real_datasets_basic_numerics}
\begin{figure}%
    \centering
    \includegraphics[width=0.9\linewidth]{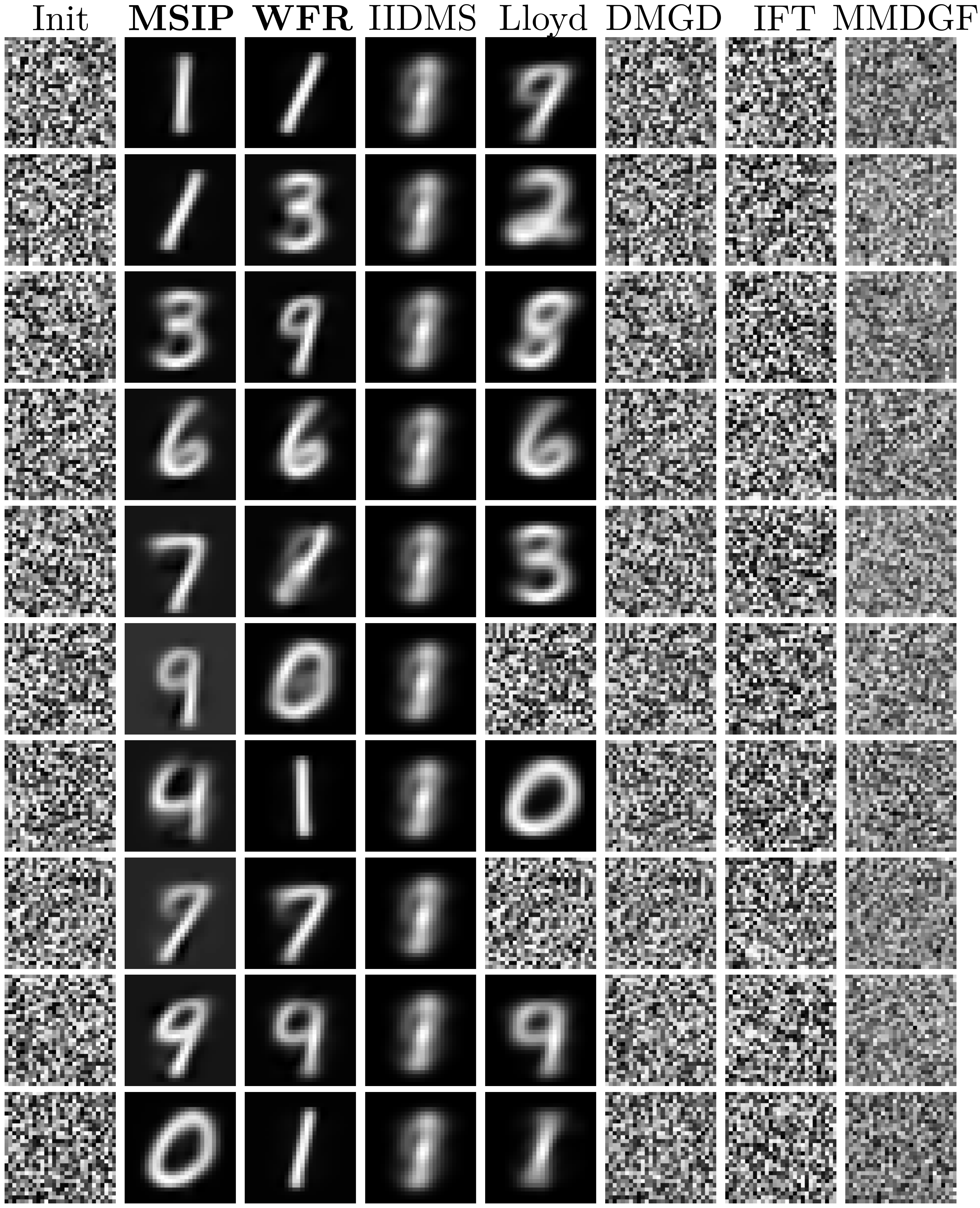}
    \caption{Comparing quantizations of MNIST}
    \label{fig:mnist_comparison}
\end{figure}
We now illustrate our algorithms using the MNIST dataset~\citep{lecun2010mnist}; for further results, see \Cref{quant:mnist}. We compare MSIP, Lloyd's algorithm, WFR, IFTflow, MMDGF, DMGD, and classical (non-interacting) mean shift (IIDMS). When Lloyd's algorithm produces an empty Voronoi cell, we make the corresponding particle retain its position. 

\Cref{fig:mnist_comparison} shows the results of $5000$ iterations for each algorithm using a Mat\'{e}rn kernel, all initialized with the same set of $M$ i.i.d.\ samples drawn uniformly from $[0,1]^{784}$. %
We see that MSIP and WFR recover recognizable digits. MSIP and WFR each output two particles that look like ones: these image pairs correspond to different modes of MNIST, distant in pixel-space. %
IIDMS collapses to a single mode, showing no diversity whatsoever. Lloyd's algorithm has difficulty  using all centroids, as several particles have empty Voronoi cells across all iterations%
\footnote{Lloyd's algorithm is typically initialized from the data distribution itself to mitigate this phenomenon.}; %
the cells of the remaining particles cover many digits. The other three algorithms remain trapped far from the distribution's support. %
This experiment illustrates the mode-seeking behavior of our algorithms for multi-modal distributions and initialization robustness in high-dimensional domains.

\section{DISCUSSION}\label{sec:discussion}

This work suggests further theoretical developments for MMD minimization that bypass mean-field analysis. In particular, the unifying perspective developed here connects, for the first time, MMD minimization to %
convergence theory for Lloyd’s algorithm and the mean shift algorithm. We expect that this connection will provide new tools for analyzing the difficult non-convex optimization problems intrinsic to quantization.
We have already characterized, e.g., via \cref{prop:wfr_convergence} and \cref{thm:gradient_opt_mmd}, non-trivial properties of WFR-IPS and MSIP. Future work
could investigate possible {\L}ojasiewicz properties for our algorithms; indeed, Kurdyka--{\L}ojasiewicz inequalities were  recently used to establish theoretical guarantees for both Lloyd's algorithm and mean shift \citep{PoCaPa24,YaTa24}. %
Beyond convergence guarantees, an interesting direction for future research is to explore the impact of weighted quantization on the understanding of neural network training~\citep{ArKoSaGr19,RoJeBrVa19} and on sparse measure reconstructions~\citep{DeGa12,Chi22,BeGr24}. We also believe that MSIP and WFR-IPS can be extended for sampling from unnormalized densities \citep{LiWa16,ChMaGoBrOa18,KoAuMaAb21,MaMa24}. Generally, our methods show promising results for mode seeking using interacting particle systems, %
indicating that such systems are well suited to general problems %
of non-convex high-dimensional distribution approximation.

\subsubsection*{Acknowledgments}
{AB,} DS, and YM acknowledge support from the US Department of Energy (DOE), Office of Advanced Scientific Computing Research, under grants DE-SC0021226 (FASTMath SciDAC Institute) and DE-SC0023188. AB and YM also acknowledge support from the ExxonMobil Technology and Engineering Company. DS also acknowledges support from a 2025--26 MathWorks Engineering Fellowship at MIT.
DS and YMM  acknowledge support from the U.S. Department of Energy, Office of Science, Office of Advanced Scientific Computing Research, via the M2dt MMICC center under award number DE-SC0023187.
The authors also acknowledge the MIT Office of Research Computing and Data, MIT SuperCloud, and Lincoln Laboratory Supercomputing Center for providing HPC resources that have contributed to the research results reported within this paper.

\bibliography{references}

\clearpage
\appendix

\onecolumn

\clearpage
\section{Additional numerical details}\label{sec:settings_numerics}

In this section, we provide details on the algorithms, numerical simulations presented in \Cref{sec:numerics}, along with additional simulations. 

\subsection{Algorithm listings}\label{sec:alg_lists}

\RestyleAlgo{ruled}
\SetKwInput{KwData}{Input}
\SetKwInput{KwResult}{Output}
\DontPrintSemicolon

\begin{algorithm}
\caption{Wasserstein--Fisher--Rao Interacting Particle System for MMD: Euler discretization}
\label{alg:wfr_gf}
\KwData{Samples $x_1,x_2,\ldots,x_N\sim\pi$, optimization iterations $T$, differentiable kernel $\kappa:\X\times\X\to\RR^+$, step size $\eta$, speed $\alpha\in\RR^+$, original configuration $y^{(0)}_1,y^{(0)}_2,\ldots,y^{(0)}_M\in\X$ and weights $w^{(0)}_1,w^{(0)}_2,\ldots,w^{(0)}_M\in\RR$}
\KwResult{Centroids $y_1^{(T)},\ldots,y_M^{(T)}\in\X$, weights $w_1^{(T)},\ldots,w_M^{(T)}\in\RR$}
Set $\vzero(y) = \dfrac{1}{N}\sum\limits_{j=1}^N \kappa(x_j,y)$\;
\For{$t = 0,\,1,\,\ldots,\,T-1$}{
    \For{$i=1,\,2,\,\ldots,\,M$}{
        $\yitp = \yit - \eta\,\alpha\left(\sum\limits_{m=1}^M \wmst\nabla_2\kappa(\ymst,\yit) - \nabla \vzero(\yit)\right)$\;
        $\witp = \wit - \eta\, \wit\left(\sum\limits_{m=1}^M \wmst \kappa(\ymst, \yit) - \vzero(\yit)\right)$\;
    }
}
\end{algorithm}

\begin{algorithm}
\caption{Mean shift interacting particles (MSIP)}
\label{alg:msip}
\KwData{Samples $x_1,x_2,\ldots,x_N\sim\pi$, optimization iterations $T$, differentiable kernel $\kappa:\X\times\X\to\RR^+$, kernel $\bar{\kappa}$ s.t. $\nabla_2 \kappa(x,y) = (x-y)\bar{\kappa}(x,y)$, step size $\eta$, original configuration $y^{(0)}_1,y^{(0)}_2,\ldots,y^{(0)}_M\in\X$}
Set $v_0(y) = \dfrac{1}{N}\sum\limits_{j=1}^N \kappa(x_j,y)$\;
\For{$t=0,\,1,\,\ldots,\,T-1$}{
    Calculate $\bm{K}(\yt)$ and $\bar{\bm{K}}(\yt)$ as $(\bm{K}(\yt))_{i,m} = \kappa(\yit,\ymst)$ and $(\bar{\bm{K}}(\yt))_{i,m} = \bar{\kappa}(\yit,\ymst)$\;
    Set $\vzerov(\yt)$ as $(\vzerov(\yt))_i = \vzero(\yit)$ and $\whatt = \bm{K}(\yt)^{-1}\vzerov(\yt)$\;
    Set $\bmWt = \mathrm{diag}(\whatt)$ and $\hvonem(\yt)$ such that $(\hvonem(\yt))_{i,:} = \nabla \vzero(\yit) + \yit\sum_{m=1}^M\whatmst\bar{\kappa}(\ymst,\yit)$\;
    Calculate $\ytplus = (1-\eta)\yt + \eta\bmWt^{-1}\bar{\bm{K}}(\yt)^{-1}\hvonem(\yt)$\;
}
Calculate $\hat{\bm{w}}(\y^{(T)}) = \bm{K}(\y^{(T)})^{-1}\vzerov(\y^{(T)})$ with $\bm{K}(\y^{(T)})$ and $\vzerov(\y^{(T)})$ as above\;
\end{algorithm}

\subsection{Implementation Details}\label{app:implementation_details}
First, we describe the experiment in \cref{fig:intro}.
\begin{itemize}
    \item We use ten particles drawn i.i.d.\ from a Gaussian over the top-right mode with mean $(0.98,0.68)$ and covariance $\sigma^2I$, and $\sigma^2 = 0.025$.
    \item WFR-IPS, MSIP, and mean shift use a squared exponential kernel with bandwidth 0.6.
    \item MSIP takes 1000 steps of size 0.5.
    \item WFR-IPS uses diffusion speed $\alpha=50$ and simulates using an adaptive second order Runge--Kutta solver with third order error correction~\cite{scipy20} up to time $T=2000$.
    \item Lloyd's algorithm and IIDMS use 1000 steps.
\end{itemize}
Now, we describe the details of \cref{sec:synthetic_basic_numerics}.
\begin{itemize}
    \item As a benchmark for optimal quantization with $M$ points, we use the square root of the $(M+1)$-st eigenvalue of the integration operator associated with the kernel $\kappa$ and the probability measure $\pi$, divided by $\|\vzerov\|_{\mathcal{H}}$. The eigenvalues of this operator are classically used in measuring the complexity of numerical integration in an RKHS; see \cite{Pin12,Bac17}.
    \item All experiments were performed in Python using Numba~\citep{LaPiSe15}.
    \item Computationally, we use a parallelized CPU implementation of the algorithms. These were run on a cluster using up to 32 tasks per simulation. Many of the smaller visualizations were produced from experiments on a local machine with 16 CPU cores and 32 GB.
    \item For WFR, we use a fixed step-size Runge--Kutta fourth order ODE solver for simulations. As this requires four evaluations of $\vzerov$ and $\bvonem$ for each iteration, we only run the ODE for one fourth as many iterations as the other algorithms. Outside of this experiment setting, we suggest that common \textit{adaptive} ODE solvers should provide better results. We also choose the diffusion rate $\alpha$ according to na\"{i}ve hyperparameter tuning, where simple experiments provide good results for about $\alpha=25$. 
    \item For MSIP, we use the damping factor $\eta = 0.8$ for the fixed-point iterations. Moreover, we use standard floating-point methods to ensure stability when working with terms that take extremely small values such as the functions $\vzerov$ and $\bvonem$, which we employ for the squared exponential kernel.  
    \item For MMDGF, we use a step-size of $10^{-1}$ and inject noise at each iteration via $\varepsilon_t\sim\mathcal{N}(0, \beta/\sqrt{t}),$ where $t$ is the iteration and $\beta=0.05$. This noise injection is recommended in~\cite{ArKoSaGr19} to improve convergence rates and guarantees.
    \item For IFTFlow, we chose to implement the algorithm described in \citet[Appendix A]{GlDvMiZh24} with a regularization on the weight parameter of $10^{-3}$ and, similar to \cite{GlDvMiZh24}, we perform the MMD geometry minimization via a single step of projected gradient descent.
    \item We observe that the methods proposed in both \citet{ArKoSaGr19} and \citet{GlDvMiZh24} draw new samples from the target distribution at every iteration $t$ (and hence assume access to an infinite number of samples from the target distribution); %
    here, however, we assume a fixed target distribution in each simulation and thus assume that the same $N$ samples are available across all iterations of each algorithm.
\end{itemize}

Finally, we note that, for the MNIST experiment, we use a Mat\'{e}rn kernel with a bandwidth of $\sigma = 2.25$ and smoothness of $\nu = 1.5$.

\subsection{Target distributions in \texorpdfstring{\Cref{sec:synthetic_basic_numerics}}{Section \ref{sec:synthetic_basic_numerics}}}\label{sec:settings_numerics_GMM}
\begin{figure}[!hb]
    \centering
    \includegraphics[width=0.75\linewidth]{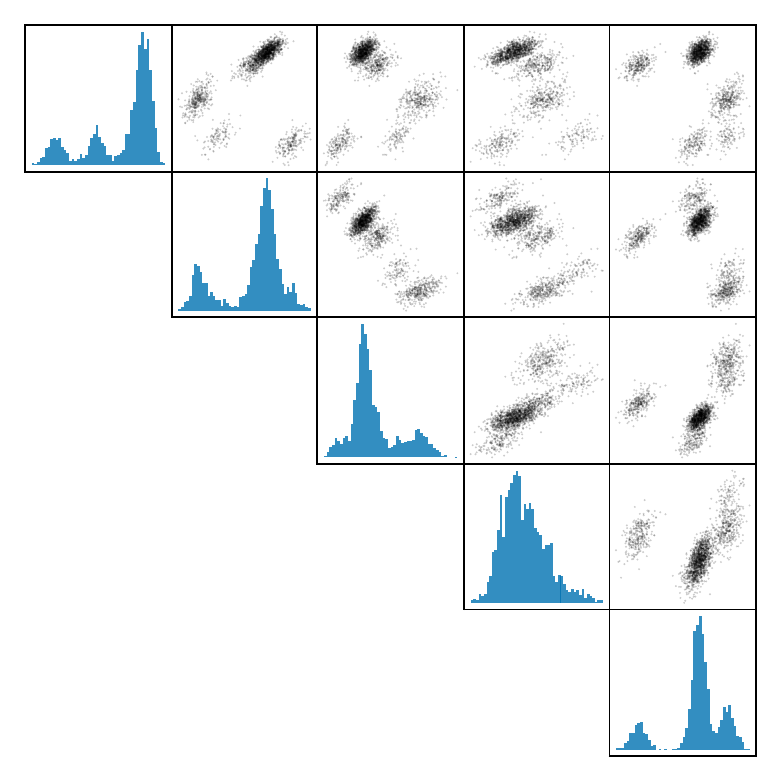}

    \caption{First five univariate and pairwise marginals of the $100$-dimensional distribution used in \Cref{sec:synthetic_basic_numerics}}
    \label{fig:gmm_marginals}
\end{figure}

We consider the following setting: $\pi$ is the empirical measure associated with $N=1000$ i.i.d.\ samples from a mixture of Gaussians $\sum_{m =1}^{L_0} \pi_{m} \mathcal{N}(c_{m}, \Sigma_{m}) $, where %
$L_0 \in \mathbb{N}^{*}$ is the number of components of the mixture, $c_1, \dots, c_{{L_{0}}} \in \mathbb{R}^{d}$ are the means of the mixtures, and $\Sigma_{1}, \dots, \Sigma_{{L_{0}}} \in \mathbb{S}^{++}_{d}$ are the covariance matrices, and $\pi_{1}, \dots, \pi_{{L_{0}}}$ are the positive weights of the mixture which satisfy $\sum_{m = 1}^{{L_{0}}} \pi_{m} =1$. We consider two versions of this mixture distribution: one in dimension $d=2$ with ${L_{0}} = 3$, and another in dimension $d = 100$ with ${L_{0}} = 5$. \Cref{fig:gmm_marginals} illustrates some marginals of the $100$-dimensional distribution used in \Cref{sec:synthetic_basic_numerics}. The 100-dimensional distribution is marginally normalized to have a covariance matrix with ones on the diagonal.

\subsection{Visualization of paths}\label{appendix:viz_paths_synthetic_dataset}

\begin{figure}[H]
    \centering
     \begin{subfigure}[t]{0.64\linewidth}
         \centering
         \includegraphics[width=0.49\linewidth]{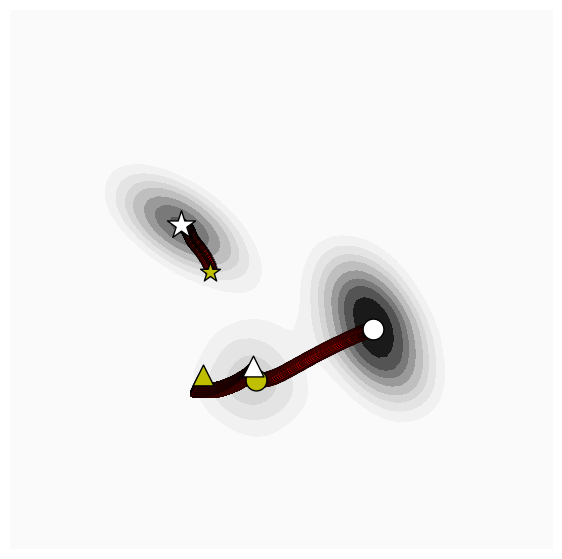}
        \hfill
        \includegraphics[width=0.49\textwidth]{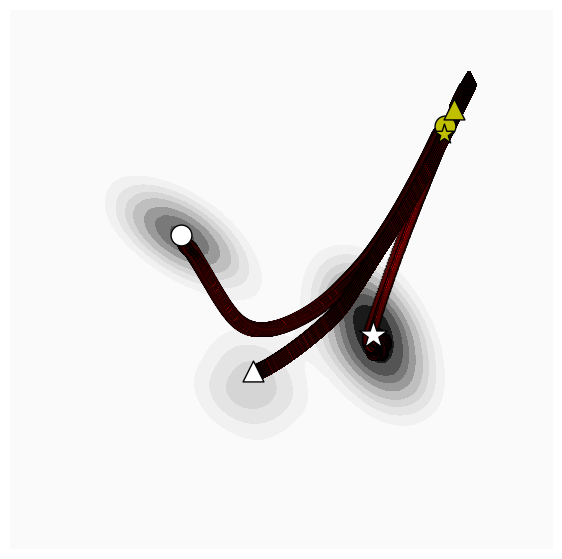}
         \caption{WFR-IPS}
     \end{subfigure}

     \begin{subfigure}[t]{0.64\linewidth}
         \centering
         \includegraphics[width=0.49\textwidth]{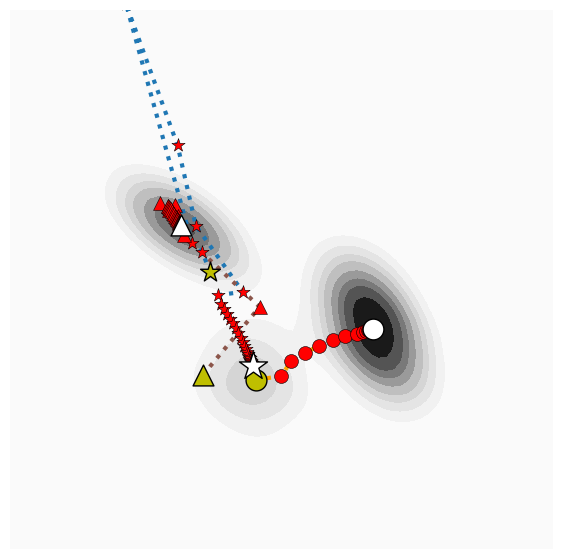}
         \hfill
         \includegraphics[width=0.49\textwidth]{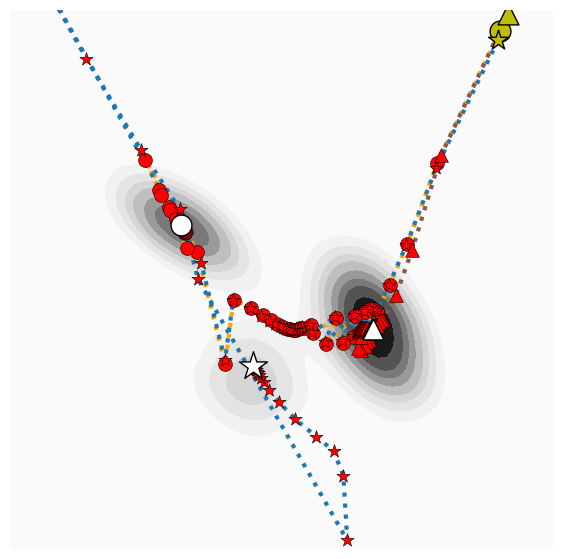}
         \caption{MSIP}
     \end{subfigure}

     \begin{subfigure}[t]{0.32\linewidth}
         \includegraphics[width=0.989\textwidth]{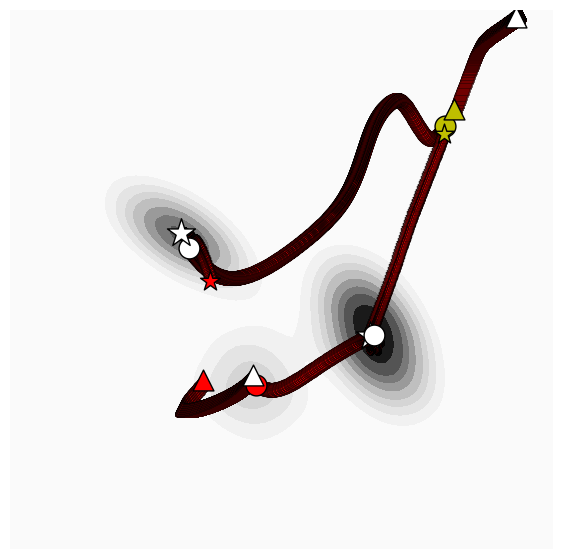}
        \caption{IFTFlow}
     \end{subfigure}
     \begin{subfigure}[t]{0.32\linewidth}
         \includegraphics[width=0.989\textwidth]{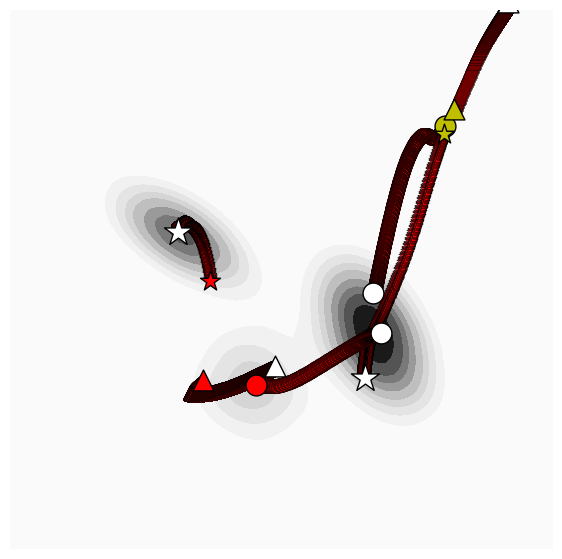}
        \caption{MMDGF}
     \end{subfigure}
        \caption{Trajectories of four algorithms started at two different intializations (yellow and red symbols). Each marker is one iteration for a particle, and the lines show  particle paths. White markers are the final particle positions.}
    \label{fig:all_trajectories}
\end{figure}

We provide a qualitative comparison of the transient behavior of the following algorithms: i) Wasserstein--Fisher--Rao Flow interacting particle system for MMD (WFR-IPS) introduced in \Cref{sec:wfr}, ii) mean shift interacting particles (MSIP) introduced in \Cref{sec:msip}, iii) IFTflow introduced in \cite{GlDvMiZh24}, iv) MMD gradient flow (MMDGF) introduced in \cite{ArKoSaGr19}. \Cref{fig:all_trajectories} compares the dynamics of the four algorithms for their first $1000$ iterations. 
For each algorithm, we illustrate the dynamics of matching $M=3$ nodes to a mixture of three Gaussians. 
For WFR-IPS and MSIP, the left and right plots in \Cref{fig:all_trajectories}(a) and (b) correspond to two different initializations: random elements of the dataset (left), and points far from the support of $\pi$ (right). For each of IFTFlow and MMDGF, we demonstrate both initializations in the same plot. The lines indicate particle trajectories, where overlapping markers indicate slower dynamics. In particular, we see WFR-IPS performing well in both scenarios with sufficient iterations. MSIP is significantly faster; 
though it produces more erratic trajectories than WFR and might first overshoot the support, it converges at the end. %
In contrast, both MMDGF and IFTFlow succeed when starting from the support of the target, but perform poorly otherwise.

\subsection{Mean shift vs.~MMD gradient flow for \texorpdfstring{$M=1$}{M=1}}\label{sec:ms_vs_ga}

\begin{figure}[!ht]
\centering
\begin{subfigure}{0.85\linewidth}
\centering
    \includegraphics[width=0.9\linewidth]{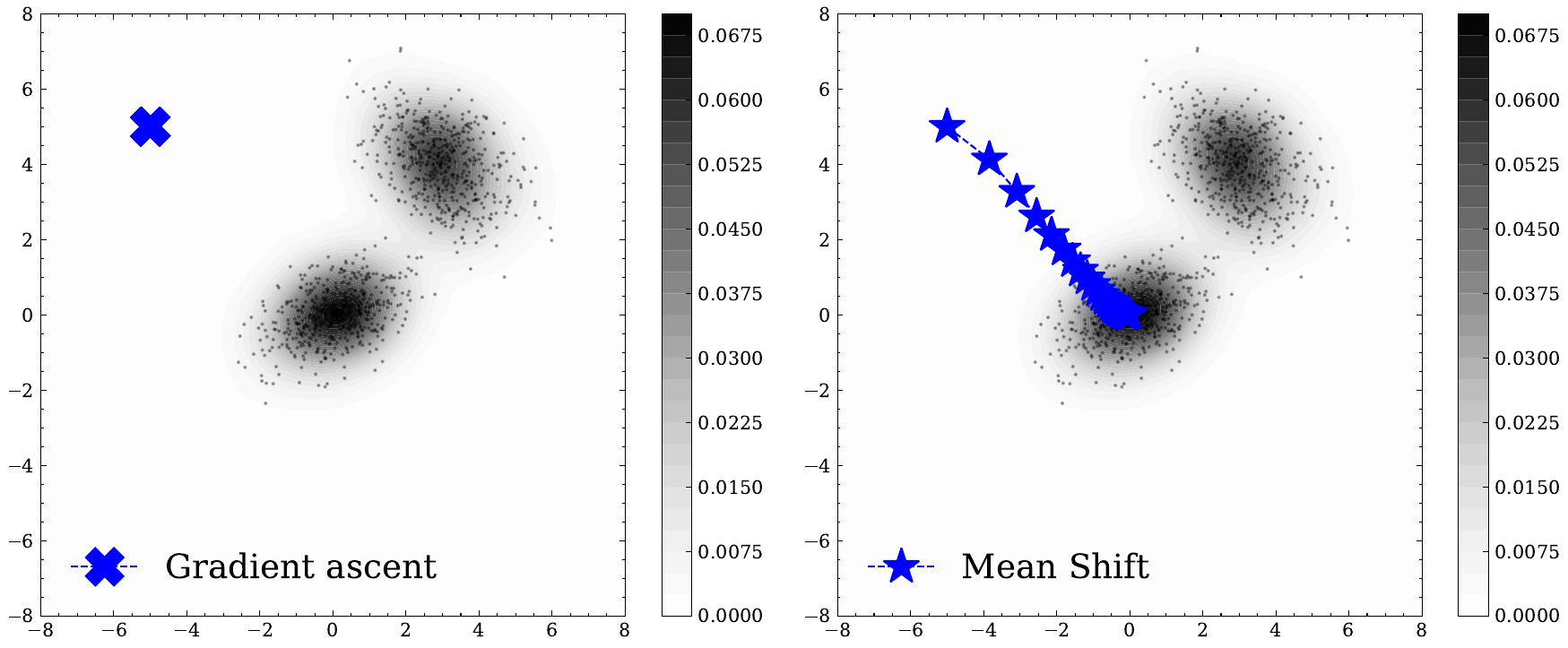}
    \caption{Dynamics of the algorithms: (left) gradient ascent on the KDE $v_0$, corresponding to MMDGF wtih $M=1$; (right) mean shift.}
    \label{fig:ms_vs_ga_a}
\end{subfigure}
     \hfill
\begin{subfigure}{0.85\linewidth}
\centering
    \includegraphics[width=0.9\linewidth]{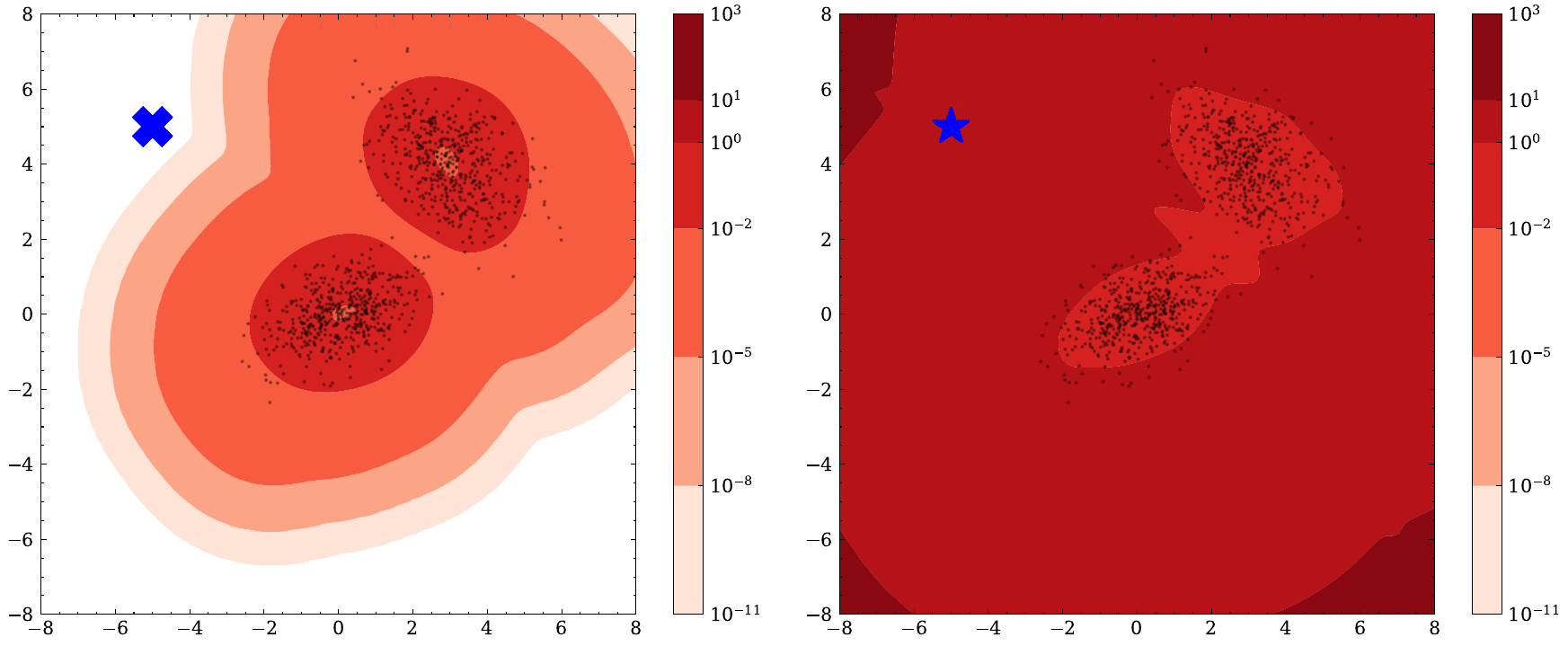}
    \caption{Norms of the descent direction vector fields of the two algorithms.  Initialization for each algorithm marked in blue. Colorbars and contour levels are identical in the left and right plots. (Left): $\|\nabla v_0\|$. (Right): $\|\nabla v_0\|/v_0$).}
    \label{fig:ms_vs_ga_b}
\end{subfigure}
     \hfill
\begin{subfigure}{0.7\linewidth}
\centering
    \includegraphics[width=0.9\linewidth]{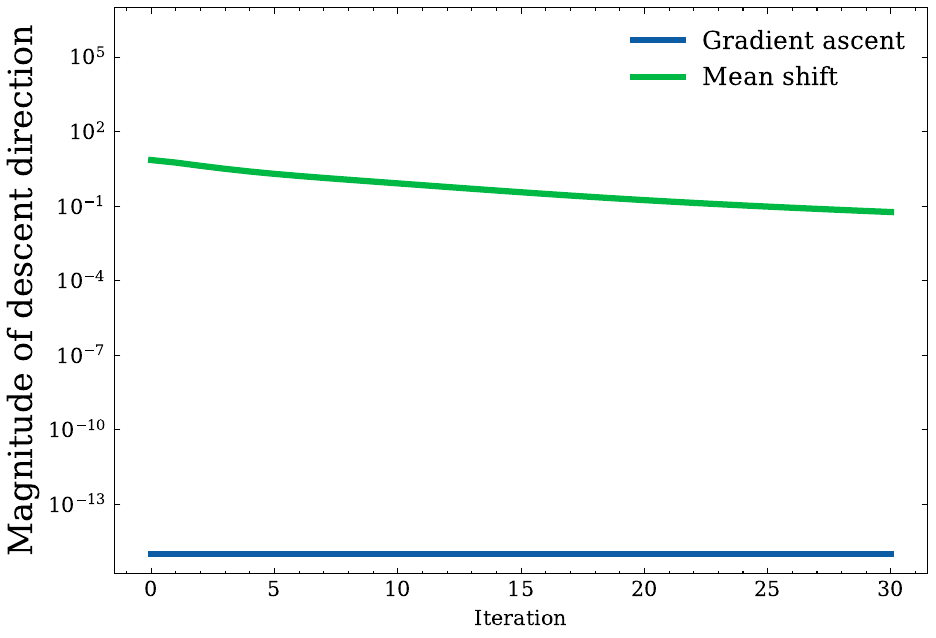}
    \caption{Comparison of the norm of the descent directions of the two algorithms across iterations.}
    \label{fig:ms_vs_ga_c}
\end{subfigure}
\caption{Comparison of the dynamics of mean shift and the discretization of MMD gradient flow proposed in \cite{ArKoSaGr19}.}
\label{fig:ms_vs_ga}
\end{figure}

\Cref{fig:ms_vs_ga_a} compares mean shift with the algorithm derived in \cite{ArKoSaGr19} for $M=1$, which we will refer to as gradient ascent on the KDE $v_0$. %
The target $\pi$ is an $N$-sample empirical representation of a mixture distribution on $\mathbb{R}^2$.
The two algorithms are initialized from the same point. We observe that mean shift converges to one of the two modes within a few iterations, whereas the particle performing gradient ascent on the KDE remains nearly stationary across iterations. The difference between the two algorithms can be explained by comparing the vector field $\nabla v_0$ to the vector field $\nabla v_0 /v_0$. The former governs the dynamics of gradient ascent; the latter governs the dynamics of mean shift. \Cref{fig:ms_vs_ga_b} compares the norms of the vector fields associated with the two algorithms. We observe that $\|\nabla v_0(y)\|$ becomes extremely small when $y$ lies far from the support of the distribution, unlike $\|\nabla v_0(y) / v_0(y)\|$, which takes significant values even  when $y$ is far from the support. This observation is further supported by \Cref{fig:ms_vs_ga_c}, which shows the evolution of the vector field norms across iterations for both algorithms.

\subsection{Additional results for the MNIST simulation}
\subsubsection{The effect of additional iterations}
\begin{figure}[!ht]
    \centering
    \includegraphics[width=0.35\linewidth]{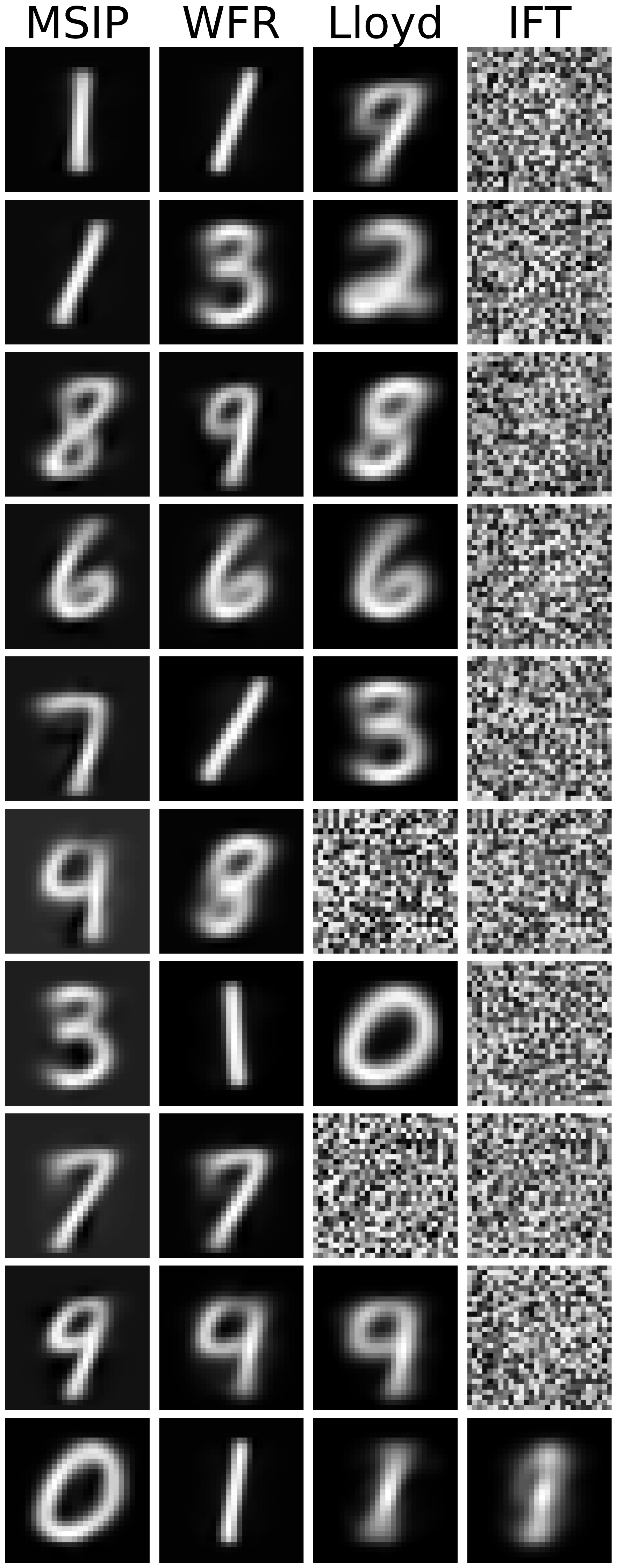}

    \caption{Comparison of four algorithms on MNIST for the iteration $T=15000$. }
    \label{fig:mnist_comparison_2}
\end{figure}

\Cref{fig:mnist_comparison_2} shows the results of the simulation detailed in \Cref{sec:real_datasets_basic_numerics} after $15000$ iterations ($10000$ more iterations than in \Cref{fig:mnist_comparison}) for MSIP, WFR-IPS, Lloyd's algorithm, and IFTFlow. We observe that their recovered images are less blurry compared to the results after $5000$ iterations. Interestingly, %
IFTFlow ends with one particle on the target's support despite not having any such particles in \cref{fig:mnist_comparison}; this particle seems to correspond visually with the mode that IIDMS finds in \Cref{fig:mnist_comparison}.

\subsubsection{Quantitative results for MNIST}\label{quant:mnist}
\begin{figure}[!ht]
    \centering
    \includegraphics[width=0.5\linewidth]{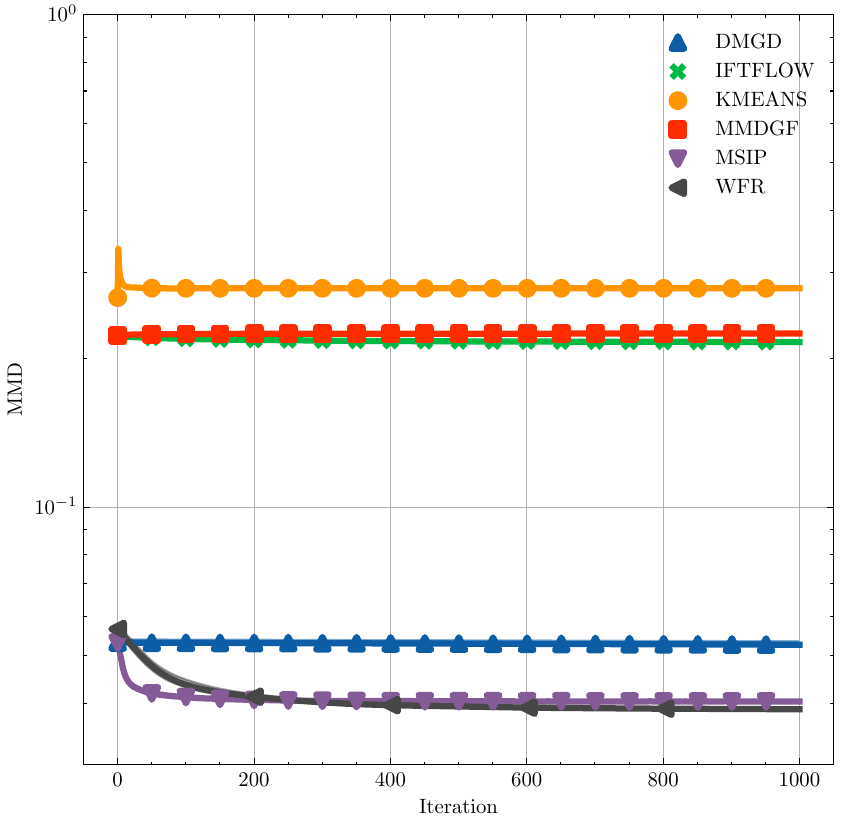}
    \caption{Comparison of different algorithms' quantization of MNIST with $M=10$ points, up to iteration $T=1000$.}
    \label{fig:mnist_comparison_quant}
\end{figure}

In \cref{fig:mnist_comparison_quant}, we evaluate the {quantitative performance} of each algorithm's quantization of MNIST. Specifically, we calculate the MMD for each algorithm using the Mat\'{e}rn kernel with $\sigma= 2.25$ and $\nu = 1.5$. We use $10$ different initializations for the experiment and show the median MMD (over these initializations) marginally at each iteration. Despite the fact that DMGD starts from a relatively low value of MMD, this value does not decrease over time for this algorithm. 
MSIP and WFR-IPS are the only two algorithms that enjoy substantial dynamics in these simulations, shown by their reduction of MMD over time.

\clearpage
\subsection{Comparison to kernel thinning}\label{sec:app_thinning}
Kernel thinning, a recent technique %
in coreset construction~\cite{DwMa24}, is another kernel-based approach to summarizing datasets. Some brief comparisons and contrasts to our approaches are as follows:
\begin{itemize}
    \item Kernel thinning methods are restricted to selecting %
    points that are in the support of the data, whereas the methods discussed here optimize point positions over the (continuous) space $\X^M$;
    \item As such, kernel thinnings are finite-time calculations, with complexity often on the order of $\mathcal{O}(N^2)$~\cite{DwMa24,DwMa21}. Neglecting $M$, %
    the methods discussed in our work have a complexity $\mathcal{O}(TN)$ for $T$ iterations; %
    whether these algorithms are faster than kernel thinning then depends on their convergence rates, e.g., if $T$ is smaller than $N$ (which is often the case when $N$ is very large), and on how the cost of each iteration scales with $M$.
    \item The number of summarization points $M$ in kernel thinning is restricted to values $M = \mathrm{floor}(N/2^r)$ for some integer $r$ due to ``halving rounds,'' whereas there are no such restrictions on the choice of $M$ for our proposed methods; 
    \item Unlike flow-based methods, kernel thinning does not involve a randomly chosen initial condition.
    Instead, stochasticity is introduced through splitting and swapping within the coreset;
    \item The standard kernel thinning discussed in~\cite{DwMa24} is \textit{unweighted}. There are algorithmic extensions to weighted coresets in this line of literature, but the construction of these weighted coresets often rely on evaluating the target score $\nabla\log\pi$.
\end{itemize}

\begin{table}[ht!]
    \centering
    \caption{Kernel thinning error quantification. $M$ is the number of nodes in our methods and $M_{kt}$ is the size of permissible coreset for kernel thinning. All error metrics are reported via the interval between the middle two quartiles (25th and 75th percentiles) of performance over 32 runs.}
    \begin{tabular}{@{} l c c c c c c @{}}\toprule
    & \multicolumn{3}{c}{Our methods} & \multicolumn{3}{c}{Kernel thinning}\\\cmidrule(lr){2-4}\cmidrule(lr){5-7}
    Example & $M$ & WFR-IPS & MSIP & $M_{kt}$ & $\MMD_u$ & $\MMD_w$ \\\midrule
         Joker & 10 & [0.164, 0.174] & [0.156, 0.159] & 9 & [0.171, 0.181] & [0.149, 0.158] \\\midrule
         GMM-2 & 3 & [0.0812, 0.144] & [0.0812, 0.0812] & 3 & [0.343, 0.346] & [0.167, 0.171]\\\midrule
         GMM-100 & 10 & [0.0493, 0.119] & [0.0282, 0.0438] & 7 & [0.342, 0.344] & [0.111, 0.112]\\
         & & & & 15 & [0.228, 0.229] & [0.0992, 0.101]\\ \bottomrule
    \end{tabular}
    \label{tab:app_kernel_thinning}
\end{table}

While the method in \cite{DwMa24} yields unweighted quantization, we also investigate a weighted variant using \eqref{eq:optimal_w}. Thus, %
we quantify the MMD of kernel thinning (KT) on a few examples in two ways: 1) the MMD on the unweighted coreset produced by KT using the same kernel we use to quantify error %
in \cref{sec:numerics} ($\MMD_u$), 2) the MMD using \textit{optimal} weights \eqref{eq:optimal_w} for result of KT ($\MMD_w$). For each example, we swap and split in KT via the same kernel we use for MSIP and WFR-IPS.%

We compare results of KT to MSIP and WFR-IPS in \cref{tab:app_kernel_thinning}, where we report our results from \cref{fig:mmd_comparison_gmm} for the GMM examples, and also show the results from each algorithm on one hundred random initializations from the joker distribution of \cref{fig:intro}. For the latter distribution, we use a step size of 0.3 for MSIP with 5000 steps, which is a very conservative choice to ensure that we avoid singularities. We simulate up to time 2000 with WFR-IPS, where $\alpha=10$, with a Runge--Kutta second order solver with third order adaptive error correction. We do not show the performance of KT on the MNIST dataset, as the time complexity of kernel thinning for such a large dataset makes the comparison unreasonable.

\subsection{Additional example configurations}\label{sec:app_examples}
\begin{figure}[ht!]
    \centering
    \includegraphics[width=0.75\textwidth]{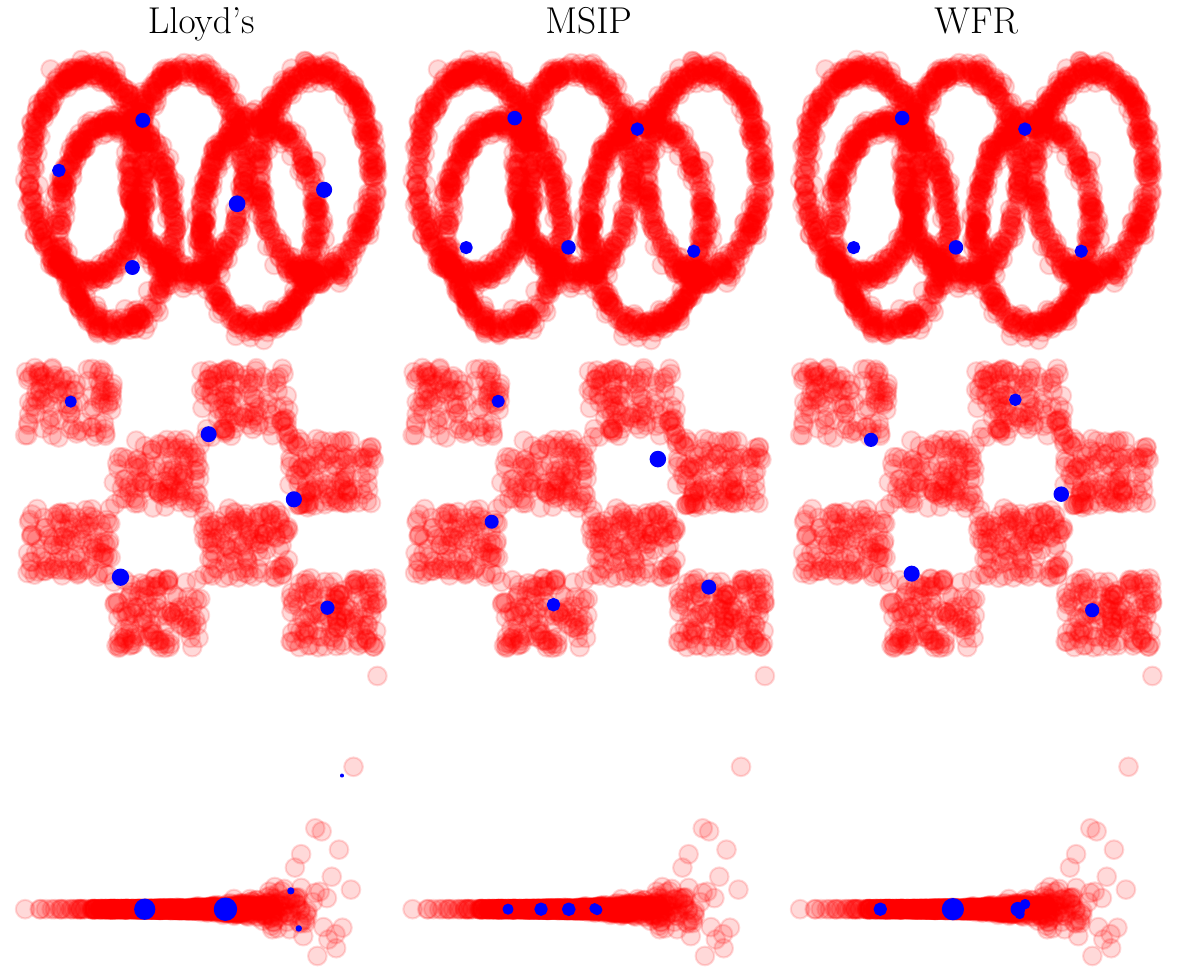}

    \includegraphics[width=0.75\textwidth]{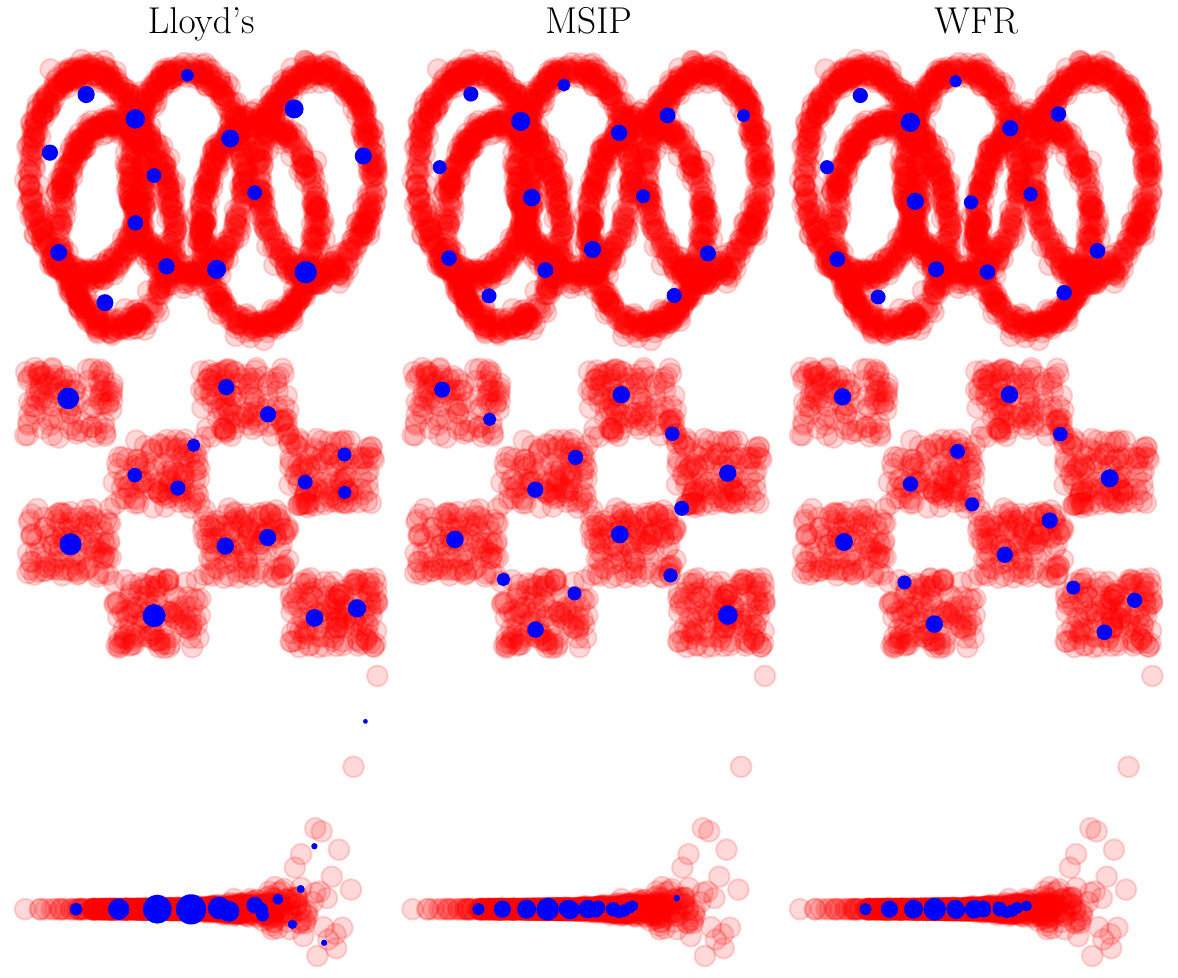}
    \caption{Final configuration of MSIP and WFR-IPS compared to Lloyd's algorithm with identical initialization for $M=5$ (top) and $M=15$ (bottom). Particle size reflects quantization weight.}
    \label{fig:app_configs}
\end{figure}

In \cref{fig:app_configs}, we show the weighted quantizations obtained using MSIP and WFR-IPS, alongside weighted quantizations from Lloyd's algorithm, for a few two-dimensional distributions with $M=5$ and $M=15$. 
The distributions are an asymmetric ``rings'' distribution (each ring $j$ has a density in polar coordinates $(r, \theta)$ given by $\pi_j(r,\theta)=U(\theta;0,2\pi)\mathcal{N}(r;1,0.05^2)$), a ``checkers'' distribution (i.e., a mixture of uniform densities $\pi(x) = \frac{1}{L_0}\sum_{j=1}^{L_0} U(x-[z^{(1)}_j,z^{(2)}_j];[0,1]^2)$), and Neal's two-dimensional funnel distribution~\cite{Nea03} (i.e., $\pi(x_1,x_2)=\mathcal{N}(x_1;0,1)\mathcal{N}(x_2;0,\exp(x_1))$).

We see that, when the number of particles is small in comparison to number of modes of the distribution, it becomes difficult to maintain nodes on the support of the distribution, particularly when the distribution has discontinuous density. This is easily explained by noting that even when the underlying $\pi$ is discontinuous, $v_0$ (its KDE) is continuous with support everywhere. Regardless, MSIP and WFR-IPS do a better job of minimizing the MMD than Lloyd's algorithm, as seen in \cref{tab:app_configs_MMD}. %
We also note that MSIP and WFR-IPS achieve similar final quantization performance in most cases (with the exception of the funnel distribution for $M=5$); again, this is to be expected, since both algorithms seek to minimize the MMD, and MSIP further seeks a stationary point of the WFR flow.
\begin{table}[ht!]
    \centering
    \caption{MMD values for configurations shown in \cref{fig:app_configs}.}
    \label{tab:app_configs_MMD}
    \begin{tabular}{@{} c c c c c c c@{}}\toprule
         & \multicolumn{2}{c}{Rings} & \multicolumn{2}{c}{Checkers} & \multicolumn{2}{c}{Funnel}\\\cmidrule(lr){2-3}\cmidrule(lr){4-5}\cmidrule(lr){6-7}
        &  $M = 5$ & $M = 15$ & $M = 5$ & $M=15$ & $M=5$ & $M=15$\\\midrule
        Bandwidth & 0.8 & 0.4 & 1.0 & 0.4 & 4.0 & 0.8\\\midrule
         MSIP & \textbf{0.132} & \textbf{0.0815} & 0.0947 & \textbf{0.0680} & 0.113 & 0.0466\\
         WFR &\textbf{0.132} & 0.0830 & \textbf{0.0930} & 0.0682 & \textbf{0.0257} & \textbf{0.0441}\\
         Lloyd's & 0.203 & 0.122 & 0.143 & 0.116 & 0.140 & 0.184
    \\\bottomrule
    \end{tabular}
\end{table}

\clearpage

\subsection{Weight positivity}\label{sec:app_weights}
\begin{figure}[ht!]
    \centering
    \includegraphics[width=0.9\linewidth]{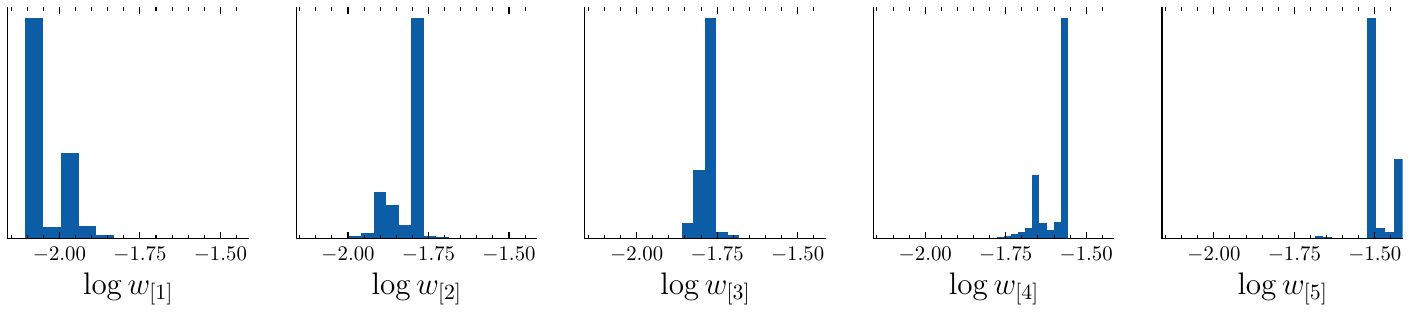}
    
    \includegraphics[width=0.9\linewidth]{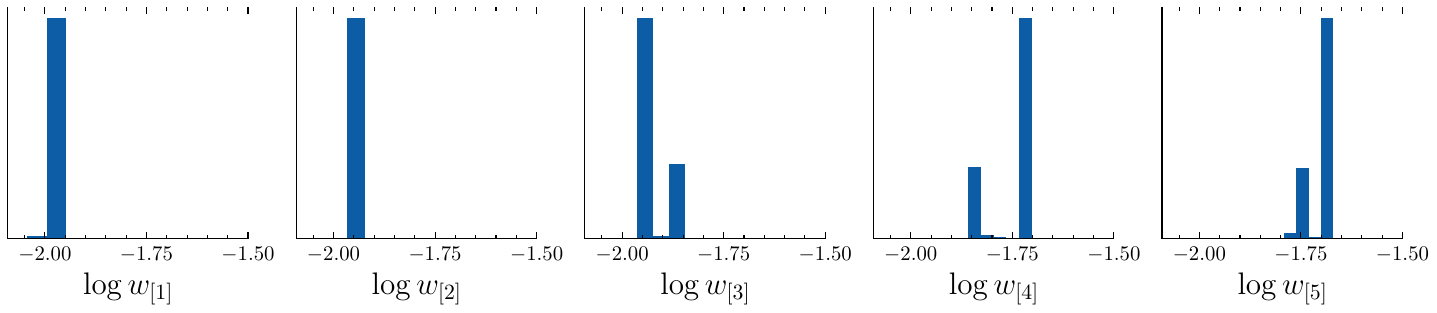}
    \caption{Weights of WFR-IPS trajectories, marginalizing out time: The weights are sorted at each timestep, and the distribution across time is plotted above, where weights increase left to right (ordering statistic subscript $[j]$ is the $j$th smallest). (Top): Checkers target. (Bottom): Rings target.}
    \label{fig:app_wfr_wts}
\end{figure}

\begin{figure}[ht!]
    \centering
    \includegraphics[width=0.9\linewidth]{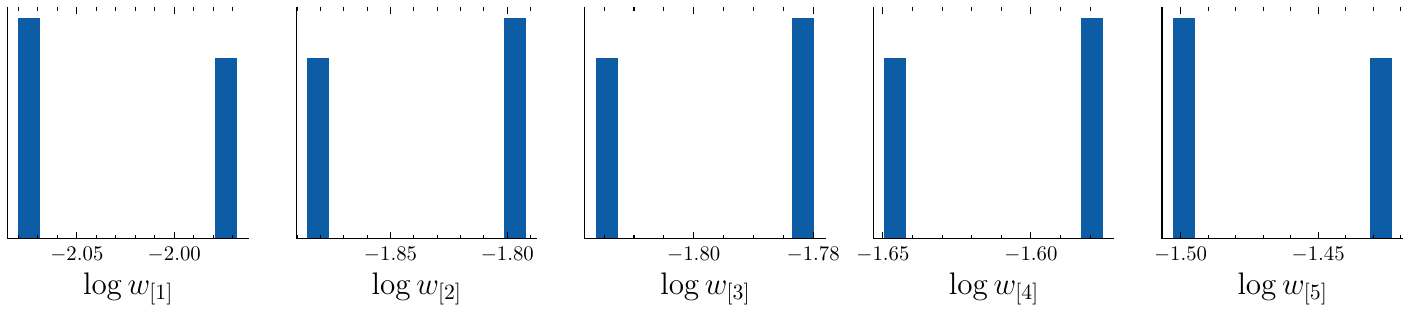}
    \includegraphics[width=0.9\linewidth]{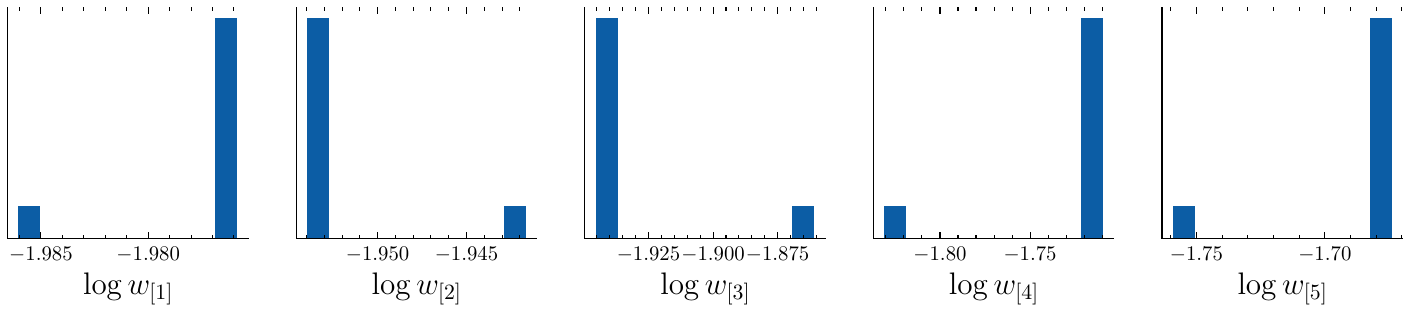}
    \caption{Weights of MSIP \textit{final} configurations. The weights increase from left to right (ordering statistic subscript $[j]$ is the $j$th smallest). (Top): Checkers target. (Bottom): Rings target.}
    \label{fig:app_msip_wts}
\end{figure}

We see, using logarithmic horizontal scaling in \cref{fig:app_wfr_wts}, that WFR-IPS in practice produces weights in $(0,1)^d$ across twenty runs for the checkers and rings distributions. These two-dimensional examples are run with a Runge--Kutta second order integrator with third-order error correction. Similarly, we examine the MSIP configurations at the \textit{final} time in \cref{fig:app_msip_wts}. %
We observe \textbf{positivity of the final weights across all one hundred random initializations}. Over all one hundred \textit{entire} trajectories for all $M=5$ weights, only 0.0628\% of weights are nonpositive in the checkers example and 0.0472\% in the rings example. These results indicate that, while negative weights may happen in MSIP iterations, they are not observed near the steady state, nor are they common in the transient state.

\section{Application to cell photography dataset}\label{app:dataset}
\begin{figure}[t!]
    \centering
    \includegraphics[width=0.4\linewidth]{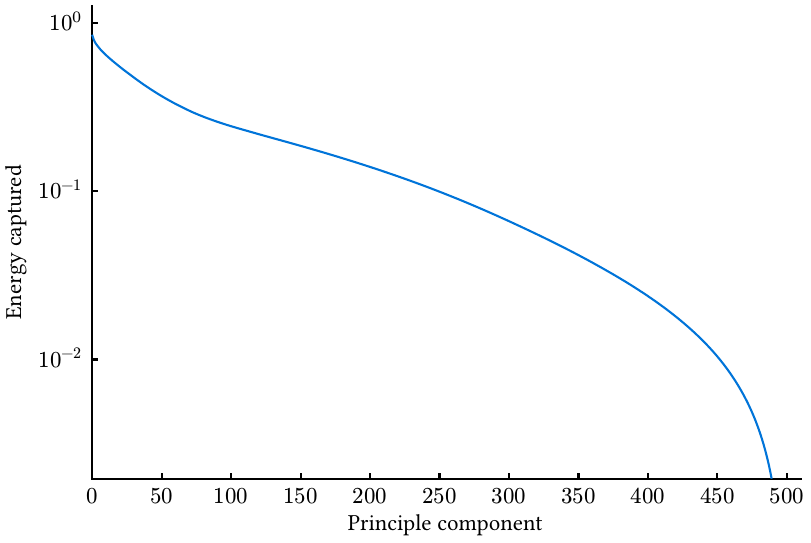}
    \caption{Residual squared energy from first 500 principle components. Given the left singular vectors $u_i$ and singular values $\sigma_i$ of the centered samples, the curve $f$ is defined as $f_i = \frac{\sum_{j=i+1}^{500}\sigma_j^2}{\sum_{j=1}^{500}\sigma_j^2}$.}
    \label{fig:dataset_vals}
\end{figure}

\begin{figure}[htbp!]
    \centering
    \includegraphics[width=0.5\linewidth]{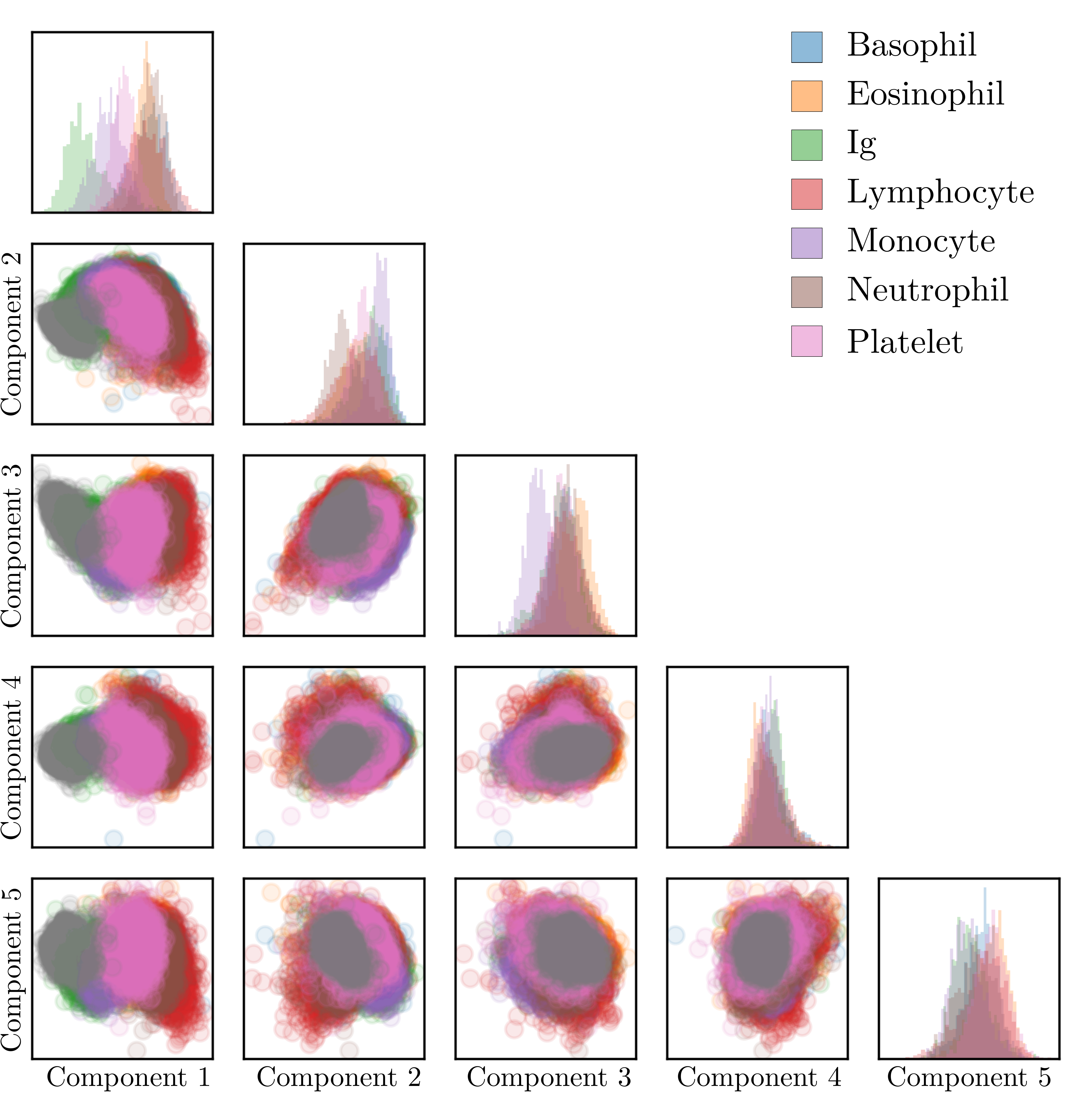}
    \caption{First five principle components of the dataset in \Cref{app:dataset}, colored by the true label of the sample.}
    \label{fig:dataset_pca_groups}
\end{figure}

We use a dataset furnished by~\cite{AcAlMePuRo19}, which consists of photographs for seven different types of cells in a bloodstream, where each photograph is $360 \times 363$ pixels (i.e. a true dimension of 131040 per sample for each color channel). We then convert these photos to grayscale and reduce the dimensionality of the photographs using principle component analysis (PCA). While this is not sufficient for reproducing the data (the algebraic decay of the first 500 singular values can be seen in \cref{fig:dataset_vals}),
it is adequate for clustering, enabling us to distinguish between cell types and capture intra-class variability.
In \cref{fig:dataset_pca_groups}, we visualize the different cell groupings when projected onto the first five principle components---note that there is natural groupings of the images, which may be overlapping when viewing low-dimensional projections. In this example, we keep the first 100 principle components then run Lloyd's algorithm and MSIP each for 500 iterations. We initialize each with 21 samples from the dataset, where MSIP uses a squared-exponential kernel with bandwidth 0.025.

We note that both Lloyd's algorithm and MSIP finish 500 iterations on this $d=100$ dimensional dataset over 16339 samples for 21 centroids in less than ten seconds of wall-clock time.\footnote{This was tested on an Apple 12-core M2 Pro processor using the provided code.} This clearly shows practicality in the low-centroid regime%
. Further, this can be accelerated greatly by more efficient implementations in the future.

\begin{figure}[h!]
    \centering
    \includegraphics[width=0.4\linewidth]{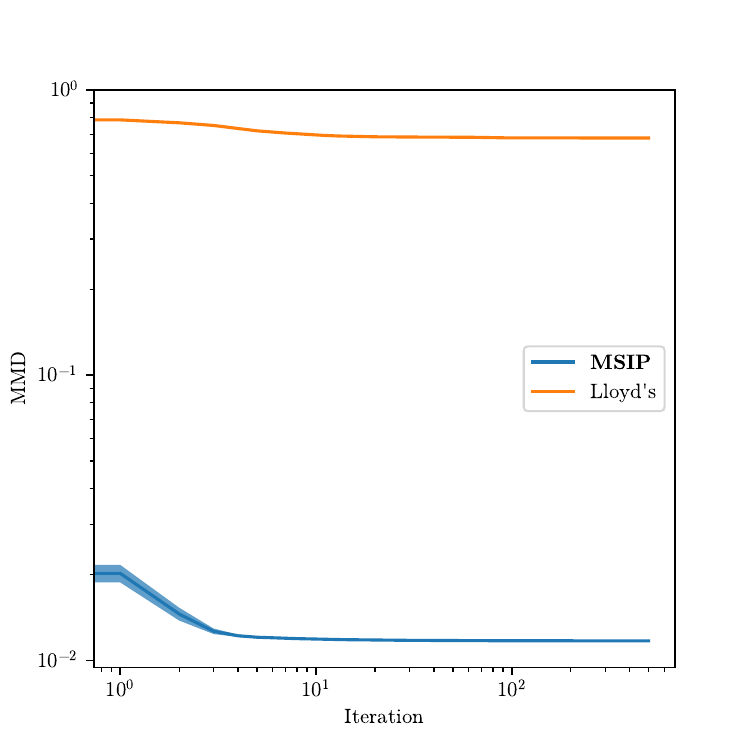}
    \caption{Convergence of the MMD for each of Lloyd's and MSIP over ten independent trials, where each algorithm is initialized identically to the other for each trial. MMD measured with bandwidth 0.025.}
    \label{fig:dataset_mmd_conv}
\end{figure}

\Cref{fig:dataset_mmd_conv} shows the evolution of the MMD for the two algorithms over the iterations. We observe that MSIP outperforms Lloyd's regardless of the initialization, which is expected by the construction of the method. \Cref{fig:dataset_centroids} shows the dataset projected onto its first two principal directions, along with the projections of the particles obtained by both algorithms. We observe that both algorithms yield configurations covering the distribution in some sense. Now, we compare the intra-centroid distances for each of the two algorithms to gauge diversity of the particles. MSIP achieves a minimum intra-centroid distance of $0.0295$ for the example in \cref{fig:dataset_centroids}, while Lloyd's algorithm only $0.0249$. On the other hand, the largest intra-centroid distance for MSIP is $0.0439$, compared to $0.0422$ for Lloyd's algorithm. In other words, the configuration obtained by MSIP has higher variance while capturing the distribution better (in the MMD).

\begin{figure}[t!]
    \centering
    \includegraphics[width=0.9\linewidth]{figs/dataset/compare_example.pdf}
    \caption{Visualizing the centroids obtained for \cref{app:dataset} given a fixed initialization. Left: MSIP. Right: Lloyd's. We overlay the particle marginal locations onto the marginal histograms as vertical lines, and use black points (whose radius corresponds to the weight of the particle) for the two-dimensional marginals. The axes and remaining colors are identical to those in \cref{fig:dataset_pca_groups}.}
    \label{fig:dataset_centroids}
\end{figure}

\clearpage
\section{Existing work on gradient flows for MMD minimization}\label{sec:appendix_gf_mmd}
In the following, we review a few constructions of gradient flows for different choices of metric $D$ to minimize the $\MMD$ functional
\begin{equation}%
    \F[\mu] := \frac{1}{2} \MMD(\mu, \pi)^2.
\end{equation}
When the metric $D$ corresponds to the $\MMD$,\footnote{Again, not to be confused with when $\F$ is the $\MMD$.} the equation \eqref{eq:abstract_gradient_flow} simplifies to $\dmut = \mu_{t} - \pi$, whose solution is given by $\mut = e^{-t} \mu_{0} + (1-e^{-t}) \pi$, where $\mu_{0}$ is the initial condition \citep[Theorem 3.4]{GlDvMiZh24}. %
This solution is not suitable for the quantization problem we consider, as $\pi$ generally cannot be expressed as a mixture of $M$ Diracs.

\citet{ArKoSaGr19} studied the gradient flow of $\F$ under the 2-Wasserstein ($W_2$) geometry, corresponding to the solution of
\begin{equation}\label{eq:MMD_gf_equation}
    \dmut = \mathrm{div} \left( \mut\nabla \Ffv [\mut] \right),
\end{equation}
where $\Ffv[\mu]$ is the first variation of $\F$ at $\mu$ given by
\begin{equation}\label{eq:MMD_first_variation}
\Ffv[\mu](\cdot) = \int_{\X} \kappa(\cdot,x)\,\mathrm{d}\mu(x) - \int_{\X} \kappa(\cdot,x)\dpix.
\end{equation}

The authors propose various numerical solutions to \eqref{eq:MMD_gf_equation} based on discretizations in time and space. In particular, they replace the measures $\mut$ by empirical counterparts, yielding a system of interacting particles. Moreover, they derive conditions for the convergence of their numerical schemes, and study the importance of \emph{noise injection} as a regularization method.
However, the approaches deteriorate dramatically when initialized poorly, requiring a large number of particles to converge. Further, rigidly constraining particles to have uniform weights is problematic; this can be seen for $M=1$ when comparing gradient descent on the MMD to mean shift, which we discuss in \Cref{sec:msip}.

\cite{AlHeSt23} and \cite{HeWaAlHa24} also study Wasserstein gradient flow of MMD, exploiting the particular structure induced by Riesz kernels to develop more efficient algorithms in the context of neural sampling schemes.

An alternative geometry called \emph{spherical interaction-force transport} (IFT) is proposed and studied in \cite{GlDvMiZh24,ZhMi24}. Inheriting properties of both the $W_2$ and MMD geometries, the IFT gradient flow satisfies
\begin{equation}\label{eq:ift_gf_equation}
     \dmut = \alpha\ \mathrm{div} \left( \mu_{t} \nabla \Ffv [\mut] \right) - \beta(\mut - \pi),
\end{equation}
where $\alpha, \beta\in\RR$ are positive. The authors prove that if $\mut$ solves \eqref{eq:ift_gf_equation}, then $\F[\mut]$ decays exponentially with $t$ \citep[Theorem 3.5]{GlDvMiZh24}.  While this is a remarkable theoretical result, it only holds in the mean-field limit $M \rightarrow +\infty$. They discretize  \eqref{eq:ift_gf_equation} using a system of weighted interacting particles; as we show in \cref{appendix:viz_paths_synthetic_dataset}, many of these particles tend to drift far away from the modes of $\pi$, making the resulting quantization inefficient for, e.g., quadrature. %
Further, the proposed implementation is complicated by an inner $M$-dimensional constrained optimization problem at every iteration.

Finally, the Wasserstein--Fisher--Rao (WFR) geometry %
allows both mass transport and total mass variation. The PDE corresponding to the WFR metric is
\begin{equation}\label{eq:WFR_for_MMD_APP}
    \dmut = \alpha\,\mathrm{div}(\mut \nabla \Ffv[\mut]) - \beta\,\Ffv[\mut] \mut.
\end{equation}
This corresponds to a gradient flow of the functional $\F$%
, extending the $W_2$ distance to measures with different total masses, i.e., \emph{unbalanced} optimal transport. See \citep{KoMoVo16,GaMo17,ChPeScVi18,LiMiSa18} for details about the WFR metric, its geometric properties, and the construction of WFR gradient flows. %
Additionally, there are theoretical and numerical results for the minimization of a kernelized KL divergence as the energy (compared to our usage of the MMD) using the WFR geometry~\citep{LuLuNo19,LuSlWa23,YaWaRi24}.

Compared to \eqref{eq:MMD_gf_equation}, \cref{eq:WFR_for_MMD} introduces $\Ffv[\mu_t]\mu_t$ as a reaction term, which allows the total mass of $\mu_t$ to change as a function of $t$. %
Unbalanced transport allows for this creation or destruction of mass, which might be helpful when mass conservation is not a natural assumption, as in the case  of the minimization problem \eqref{eq:constrained_MMD_minimization}. Indeed, the weights in this optimization problem do not necessarily lie on the simplex, so the resulting quantization is not, in general, a probability measure. %
Despite this formulation, the numerical scheme proposed by \cite{GlDvMiZh24} to solve \eqref{eq:WFR_for_MMD} artificially conserves mass by projecting the weights onto the simplex; this is inconsistent with the PDE's theoretical properties.%

\section{Proofs}\label{app:proofs}

\subsection{Proof of \texorpdfstring{\Cref{prop:coupled_ode_WFR}}{Theorem \ref{prop:coupled_ode_WFR}}} \label{proof:coupled_ode_WFR}

This is a ready extension to common proof techniques in, e.g., \citep{YaWaRi24} and \citep[Section 5.7.]{ChNiRi24}. Let $\varphi \in C_{c}^{\infty}$ be a test function. We have
\begin{align*}
    \frac{\mathrm{d}}{\mathrm{d}t} \int_{\mathbb{R}^{d}} \varphi(x) \mu_{t}(\mathrm{d}x) & = \frac{\mathrm{d}}{\mathrm{d}t} \left[ \sum\limits_{i = 1}^{M} \wit  \varphi(\yit) \right] \\
    & = \sum\limits_{i = 1}^{M} \frac{\mathrm{d}}{\mathrm{d}t} \wit  \varphi(\yit) + \sum\limits_{i = 1}^{M}\wit \frac{\mathrm{d}}{\mathrm{d}t} \varphi(\yit)   \\
    & = \sum\limits_{i = 1}^{M} \frac{\mathrm{d}}{\mathrm{d}t} \wit  \varphi(\yit) + \sum\limits_{i = 1}^{M}\wit \langle \nabla \varphi(\yit), \frac{\mathrm{d}}{\mathrm{d}t} \yit \rangle    \\
    & = - \sum\limits_{i = 1}^{M} \Ffv[\mu_t](\yit) \wit  \varphi(\yit) - \alpha\sum\limits_{i = 1}^{M} \wit \left\langle \nabla \varphi(\yit), \nabla \Ffv[\mu_t](\yit) \right\rangle    \\
    & = -  \int_{\mathbb{R}^{d}} \Ffv[\mu_t](x) \varphi(x)  \mu_{t}(\mathrm{d}x) -\alpha 
    \int_{
    \mathbb{R}^d} \left\langle \nabla \varphi(x), \nabla \Ffv[\mu_t](x) \right\rangle \mu_{t}(\mathrm{d}x)\\
    & = -  \int_{\mathbb{R}^{d}} \varphi(x) \Ffv[\mu_t](x)   \mu_{t}(\mathrm{d}x) + \alpha
    \int_{
    \mathbb{R}^d}  \varphi(x) \mathrm{div}(\mu_t \nabla \Ffv[\mu_t]  )(x) \mathrm{d}x.
\end{align*}

In other words, $(\mu_{t})_{t\geq 0}$ satisfies \eqref{eq:WFR_for_MMD} in the sense of distributions.

\subsection{Proof of \texorpdfstring{\Cref{prop:wfr_convergence}}{Proposition \ref{prop:wfr_convergence}}}\label{proof:wfr_convergence}
\paragraph{Preliminary bounds}
Since $x \mapsto \kappa(x,x)$ is a constant $\kappa(x,x)\equiv B_{\kappa}$ for all $x\in\X$, we have
\begin{align}
\forall x,y \in \mathcal{X}, \:\:    \left|\kappa(x,y)\right| \nonumber 
& = \left| \langle \kappa(x,.), \kappa(y,.) \rangle_{\mathcal{H}} \right| \nonumber \\
& \leq \|\kappa(x,.)\|_{\mathcal{H}}\|\kappa(y,.)\|_{\mathcal{H}} \nonumber 
\\ &= \sqrt{\kappa(x,x) \kappa(y,y)} = B_{\kappa}.
\end{align}
In particular, we have
\begin{equation}\label{eq:positive_vzero}
\forall y \in \mathcal{X}, \:\:    |\vzero(y)| \leq \left|\int_{\mathcal{X}}\kappa(x,y) \mathrm{d}\pi(x) \right| \leq \int_{\mathcal{X}}\left|\kappa(x,y)\right|\dpix \leq B_{\kappa} \int_{\mathcal{X}}\dpix = B_\kappa.
\end{equation}
Moreover, denote by $B_{\kappa}'$ an upper bound on the gradient of the kernel $\kappa$. We have $\nabla \vzero(y) = \int_{\mathcal{X}} \nabla_{2} \kappa(x,y) \mathrm{d}\pi(x)$ for any $y \in \mathcal{X}$, so that
\begin{equation}
    \|\nabla \vzero(y) \| \leq \int_{\mathcal{X}} \|\nabla_{2} \kappa(x,y) \| \mathrm{d}\pi(x) \leq \int_{\mathcal{X}} B_{\kappa}' \mathrm{d}\pi(x) = B_{\kappa}'.
\end{equation}

\paragraph{Step 1: Existence}
The system of coupled ODEs \eqref{eq:WFR_particle_equation} can be written as $\dotz = F(z)$,
where $z \in \mathbb{R}^{M(d+1)}$ defined by the concatenation of $y_1, \dots, y_M, \bm{w}$ into an $M(d+1)-$dimensional vector, 
\begin{equation}
    z^{\top} = (y_1^{\top}, \dots, y_{M}^{\top}, w_{1}, \dots, w_{M}),
\end{equation}
and $F: \mathbb{R}^{M(d+1)} \rightarrow \mathbb{R}^{M(d+1)}$ is defined as $F(z)^{\top} = (F_{y,1}(z)^{\top}, \dots, F_{y,M}(z)^{\top}, F_{w}(z)^{\top})$, where
\begin{equation}
\forall i \in [M], \:\:  F_{y,i}(z) := -\alpha\left(\sum\limits_{m=1}^{M} w_{m} \nabla_{2}\kappa(y_m, y_i) - \nabla \vzero(y_m)\right),
\end{equation}
and 
\begin{equation}
\forall i \in [M], \:\:    F_{w}(z)_i := -w_i \Bigg( \sum\limits_{m=1}^{M} w_{m} \kappa(y_m, y_i) - \vzero(y_m) \Bigg).
\end{equation}
By assumption, $\kappa$ and $\nabla\kappa$ are continuous. Therefore, $\vzero$ and $\nabla \vzero$ are continuous and $F$ is continuous over $\mathbb{R}^{M(d+1)}$. 

By the Peano existence theorem \cite[Theorem 2.19]{Tes12}, for any $z_0 \in \mathbb{R}^{M(d+1)}$, the initial value problem

\begin{equation}
 \begin{cases}
\displaystyle \dotz &= F(z) \\
z(0) & = z_0
\end{cases}
 \end{equation}
  has at least one solution $\hat{z} : I \rightarrow \mathbb{R}^{M(d+1)}$ defined on a neighborhood $I$ of $0$ that satisfies $\hat{z}(0) = z_0$. In the following, denote $\tilde{t}$ as a positive scalar such that $[0,\tilde{t} \:] \subset I$.

\paragraph{Step 2: Positivity of the weights and universal boundedness of the weights}
Since $\hat{z}$ is continuous on $[0,\tilde{t}]$, it is bounded. In the following, we denote by $B_{\tilde{t}}>0$ that bound. We prove in the following that the corresponding weights $w_{1}^{(t)}, \dots, w_{M}^{(t)}$ are positive when $t \in [0,\tilde{t} \:]$.

For that, let $i \in [M]$, and assume that there exists $t \in (0,\tilde{t}]$ such that $\wit \leq 0$. Denote $t^{*} = \min \big \{ t \in [0,\tilde{t} \:]| \wit \leq 0 \big\}$. By definition, $\forall t \in [0,t^{*}]$, $\wit > 0$. Moreover, observe that for $i \in [M]$ and $t \in [0,\tilde{t} \:]$ we have
\begin{equation}
    \left| \sum\limits_{m=1}^{M} \wit \kappa(\ymst,\yit) - \vzero(\yit) \right|  \leq \sum_{m=1}^M |\wit|\,\kappa(\ymst,\yit) + |\vzero(\yit)|\leq (B_{\tilde{t}}M +1) B_{\kappa}.
\end{equation}
Therefore $\forall t \in [0,t^*], \:\:    \dotwit \geq -\big( (B_{\tilde{t}}M +1) B_{\kappa} + 1 \big) \wit$. Thus, using Grönwall's lemma, we prove that $\wit \geq w_{i}^{(0)} e^{-\big((B_{\tilde{t}}M +1) B_{\kappa} + 1 \big)t}$ for all $t \in [0,t^{*}]$, and as a result $w_{i}^{(t^*)}>0$, which is a contradiction. 
Therefore, for any $t \in [0,\tilde{t} \:]$,  we have $\wit > 0$.

Now, we prove that for any $i \in [M]$, we have
\begin{equation}
    \forall t \in [0,\tilde{t}\:], \:\: \wit < 1.
\end{equation}
For that, observe that by the non-negativity of the kernel ($\kappa(y_1,y_2) \geq 0$ for any $y_1,y_2 \in \mathcal{X}$ ) and the weights on $[0,\tilde{t}\:]$, we have
\begin{equation}
 \forall t \in [0, \tilde{t}\:], \:\:  \sum_{m =1}^{M} \wmst \kappa(\ymst, \yit) - \vzero(\yit) \geq B_{\kappa}\wit  - \vzero(\yit),
\end{equation}
so that 
\begin{align}
    \forall t \in [0, \tilde{t}\:], \:\: \dotwit & \leq - \wit \Big( B_{\kappa}\wit - \vzero(\yit) \Big) \nonumber \\
    & \leq - \wit \Big( B_{\kappa}\wit - B_{\kappa} \Big) \nonumber \\
    & \leq - B_{\kappa}\wit \Big( \wit - 1 \Big), \nonumber 
\end{align}
and, by defining $\tilde{w}_{i}:= 1- w_{i}$, we get $\dot{\tilde{w}}_{i} = - \dot{w}_{i}$, and
\begin{equation}
    \forall t \in [0, \tilde{t} \:], \:\:  -\dot{\tilde{w}}_{i}  \leq  B_\kappa \underbrace{(1-\tilde{w}_i)}_{w_i} \underbrace{\tilde{w}_i}_{1- w_{i}}, 
\end{equation}
so that
\begin{equation}
    \forall t \in [0, \tilde{t} \:], \:\:  \dot{\tilde{w}}_{i}  \geq  -B_\kappa(1-\tilde{w}_i) \tilde{w}_i.
\end{equation}
Now, assume that there exists $t \in (0,\tilde{t}]$ such that $\wit \geq 1$ which is equivalent to $\tilde{w}_{i}^{(t)}\leq 0$. Denote $t^{*} = \min \{t \in [0, \tilde{t} \:] | \tilde{w}_{i}^{(t)} \leq 0\}$. %
Then, for all $t\in [0,t^*]$, we know that $1 > \wit$, giving
\[(B_\kappa + 1) \geq (B_\kappa + 1)\wit.\]
By rearranging, we get
\begin{align*}
\forall t \in [0,t^{*}], \:\:    1 - \wit &\geq B_\kappa \wit - B_\kappa\\
    &=-B_\kappa(1 -\wit)\\
    \tilde{w}_i^{(t)} &\geq -B_\kappa \tilde{w}_i^{(t)}
\end{align*}

Thus, using Grönwall's lemma, we prove that $\tilde{w}_{i}^{(t)} \geq \tilde{w}_{i}^{(0)} e^{-B_{\kappa}t}$ for all $t \in [0,t^{*}]$, and as a result $\tilde{w}_{i}^{(t^{*})}>0$, which is a contradiction.

\paragraph{Step 3: Uniform boundedness of $F$}
We recall that the kernel gradient is bounded and suppose that, for any $x\in\X$ and $j\in[d]$, we have $y\mapsto \partial_{y_j}\kappa(x,y)$ is bounded by $B^\prime_\kappa$. We now show that $F:\RR^{Md}\times [0,1]^M\to \RR^{M(d+1)}$ is a bounded function. Indeed, assuming that $z\in\RR^{Md}\times [0,1]^M$, we know that
\begin{align}
    |F_{y,i}(z)_j| &= \alpha\left|\sum_{m=1}^M \wms \nabla_{2,j}\kappa(\yms,\yi) - \partial_j \vzero(\yms)\right|\\
    &\leq \alpha \sum_{m=1}^M \wms B^\prime_\kappa + B^\prime_\kappa\\
    &= 2\alpha M B^\prime_\kappa.
\end{align}
Similarly,
\begin{align}
    |F_{w,i}(z)| &= w_i\left|\sum_{m=1}^M \wms\kappa(\yms,\yi) - \vzero(\yms)\right|\\
    &\leq \sum_{m=1}^M B_\kappa + B_\kappa\\
    &= 2MB_\kappa.
\end{align}
Therefore, $F$ is uniformly bounded on the domain $\RR^{Md}\times[0,1]^M$.

\paragraph{Step 4: Extension of the solution to $[0,+\infty)$}
Suppose that we set $z_0\in\RR^{Md}\times(0,1)^M$ and consider when
\begin{equation}
 \begin{cases}
\displaystyle \dotz &= F(z) \\
z(0) & = z_0
\end{cases}
 \end{equation}
admits a maximal solution on an interval $[0,T_{\max})$ for some positive $T_{\max} < + \infty$. Then, the set 
\begin{equation}
    S_{\max}:= \overline{\{ 
    \hat{z}(t); t \in [0,T_{\mathrm{max}}) \}}
\end{equation}
is a compact subset of $\RR^{(d+1)M}$. Indeed, it is closed by construction and we have established that the weights are bounded by unity, independent of time. Since we work on a finite domain, we can bound any point
\[|y_{ij}| = \left|\int_0^{T_{\max}} F_{y,i}(z^{(t)})_j\,\mathrm{d}t\right|\leq 2\alpha M B^\prime_\kappa T_{\max}.\]
Then, we know that the set $S_{\max}$ is compact. Now, let $(t_n)_{n \in \mathbb{N}^{*}}$ be a sequence of scalars in $[0,T_{\mathrm{max}})$ such that $\lim_{n\to\infty}t_{n} =T_{\max}$. Since $S_{\max}$ is a compact set, the sequence $\hat{z}(t_n)$ has a sub-sequence $(\hat{z}(t_{n_m}))_{m \in \mathbb{N}}$ that converges to some $z_{\star}\in S_{\max}$ as $m$ goes to $+\infty$. We consider the problem 
\begin{align}
    \dot{z} & = F(z)\\
    z(0) & = z_{\star}.
\end{align}
By using Peano existence theorem again, we prove the existence of a solution defined on a neighborhood $I_{\star}$ of $0$. Thus we can extend $\hat{z}$ to  $[0,T_{\mathrm{max}} + \varepsilon_{\star})$, 
where $\varepsilon_{\star}>0$ is any positive scalar such that $[0,\varepsilon_{\star}) \subset I_{\star}$, giving a contradiction. Therefore $T_{\mathrm{max}} = +\infty.$

\paragraph{Step 5: Uniqueness of solution}
Since we have seen that $F$ is continuous, uniformly bounded on $\RR^{Md}\times(0,1)^M$, assumed that the initialization gives $w^{(0)}\in(0,1)^M$, and shown that $w^{(t)}\in(0,1)^M$ for any finite $t > 0$, we know that $F$ is a Lipschitz function on its domain and across the trajectory for all $t$ and thus, by the Picard--Lindel\"{o}f theorem, we get that the solution is unique. We first note that this theorem gives us a convergence in a small neighborhood of zero; however, we use the same argument as in step 4 by noting that the ODE is autonomous and, thus, the time-independent bound of $F$ provides the ability to stretch the uniqueness to time infinity. We then remark that we see the boundedness of $\yt$ is time-dependent in step 4. Nevertheless, having only a uniform bound on our \textit{function}, i.e., $F$, is sufficient to employ Picard--Lindel\"{o}f.

\paragraph{Step 6: MMD dissipation}
We remark that the WFR gradient flow chooses the time derivative of the functional first variation $\Ffv$ to locally minimize the Wasserstein--Fisher--Rao (or Hellinger--Kantorovich) metric~\cite{ChNiRi24},
\[D_{\mathrm{WFR}}(\psi,\mu;\alpha) = \alpha\int\|\nabla \psi\|^2\,d\mu + \int \psi^2 d\mu.\]

We can prove this using typical integration-by-parts arguments similar to, e.g., \cite{ArKoSaGr19}. Assume that $\Ffv$, the first variation of our loss functional, is sufficiently smooth and recall standard calculus of variations identities
\begin{align}
    \frac{d}{dt}\mathcal{F}[\mu_t] &= \int \left(\Ffv[\mu_t]\frac{d}{dt}\mu_t\right)\,dy\\
    &=\int \Ffv[\mu_t](\alpha\ \mathrm{div}(\mu_t\nabla \Ffv[\mu_t]) - \mu_t \Ffv[\mu_t])dy\\
    &= -\left(\alpha \int\|\nabla \Ffv[\mu_t]\|^2\,d\mu_t(y) + \int\Ffv[\mu_t]^2d\mu_t\right),
\end{align}
where the last step uses integration-by-parts. Following \cite{ArKoSaGr19}, the MMD functional $\mathcal{F}$ exhibits the regularity needed for these calculations.

\subsection{Proof of \texorpdfstring{\Cref{prop:steady_state_equation_on_y}}{Theorem \ref{prop:steady_state_equation_on_y}}}\label{proof:steady_state_equation_on_y}

Under \Cref{assumption:gradient_kappa}, for $j \in [M]$, equation \eqref{eq:MMD_gradient_mixture} can be written as
\begin{equation}\label{eq:mixture_diracs_kbar_eq_2}
   \sum\limits_{i=1}^{M} w_{i} y_{i} \bar{\kappa}(y_i,y_j) = \Big(\sum\limits_{i=1}^{M} w_{i}  \bar{\kappa}(y_i,y_j) - \int_{\mathcal{X}} \bar{\kappa}(x,y_j) \mathrm{d}\pi(x) \Big)  y_j +    \int_{\mathcal{X}} x \bar{\kappa}(x,y_j) \mathrm{d}\pi(x),
\end{equation}
which can be expressed in matrix form as \eqref{eq:mixture_diracs_kbar_eq_2_reformulation}.

\subsection{Proof of \texorpdfstring{\Cref{thm:gradient_opt_mmd}}{Theorem \ref{thm:gradient_opt_mmd}}}\label{proof:thm_gradient_opt_mmd}

First, let $C_\pi := \int_{\X} \int_{\X} \kappa(x,x^{\prime}) \dpix \mathrm{d}\pi(x^{\prime})$. Then, observe that for $\Y\in \X_{\neq}^{M}$ we have
\begin{equation}
    F_{M}(\Y) = \frac{1}{2} \bigg( C_\pi - 2 \langle \what, \vzerov(\Y) \rangle + \langle \what, \Ky \what \rangle  \bigg),
\end{equation}
where $\hat{\bm{w}}(\Y)$ is defined by \eqref{eq:optimal_w}  respectively, and $\vzero$ is given by \eqref{eq:mke_def}.

Now, using \eqref{eq:optimal_w}, we prove that
\begin{equation}\label{eq:kernel_ids}
    \langle \hat{\bm{w}}(\Y), \vzerov(\Y) \rangle = \langle \vzerov(\Y), \bm{K}(\Y)^{-1} \vzerov(\Y) \rangle = \langle \hat{\bm{w}}(\Y), \bm{K}(\Y) \hat{\bm{w}}(\Y) \rangle.
\end{equation}
In particular, the expression of $F_{M}(\Y)$ simplifies further to 
\begin{align*}
    F_{M}(\Y) & = \frac{1}{2} \bigg(C_\pi - \langle \what, \vzerov(\Y) \rangle \bigg) = \frac{1}{2} \bigg( C_\pi - \langle \what, \bm{K}(\Y) \what \rangle \bigg).
\end{align*}

In the following, we proceed to the calculation of the gradient of $F_{M}$. First, for a configuration $Y\in\X^M$, we define $y_{ij}$ as the scalar in the $j$th dimension of node $\yi$. Now, we seek the gradient $\nabla F_{M}\in\RR^{M\times d}$. In particular, for $i \in [M]$ and $j \in [d]$, we use $\nabla_{ij} F_{M}(\Y)$ to denote the $(i,j)$ entry of $\nabla F_{M}(Y)$ corresponding to input $\yij$ evaluated at $\Y\in\mathbb{R}^{M \times d}$. 
Moreover, we define $\nabla_{ij} \Ky\in\RR^{M\times M}$  to be the differentiation of each element of matrix $\bm{K}(\Y)$ with respect to the $j$th coordinate of vector $\yi$. Similarly, for kernel mean embedding $\vzerov: \mathcal{X}_{\neq}^{M} \rightarrow \mathbb{R}^{M}$ we denote by $\nabla_{ij} \vzerov$ the vector obtained by differentiating each element of $\vzerov$ with respect to the $j$th coordinate of vector $\yi$. In other words, we have
\begin{equation}
    \Dij F_{M} = \frac{\partial}{\partial \yij} F_{M}(\Y),\quad \Dij \bm{K}(\Y) = \frac{\partial}{\partial \yij} \Ky , \quad \Dij \bm{v}(\Y) = \frac{\partial}{\partial \yij} \bm{v}(\Y)
\end{equation}

Using \eqref{eq:kernel_ids} and matrix calculus, we have 
\begin{equation}\label{eq:gradient_optimal_mmd_identity}
    2\Dij F_{M}(\Y) = -2\langle \what,\,\Ky\Dij\what\rangle - \langle \what,\,\Dij\Ky\,\what \rangle .
\end{equation}
Since $\what = \Ky^{-1}\vzerov(\Y) $, we get
\begin{gather}
\Dij \what = \Ky^{-1}\Dij \vzerov(\Y) -\Ky^{-1}\,\Dij\Ky\,\Ky^{-1}\vzerov(\Y),\\
\Ky\Dij\what = \Dij\vzerov(\Y) - \Dij\Ky\,\what
\end{gather}
so that
\begin{equation}
    \langle\what,\ \Ky \nabla_{ij} \what\rangle = \langle\what,\ \nabla_{ij} \vzerov(\Y)\rangle -\langle \what,\ (\nabla_{ij}\Ky)\what\rangle.
\end{equation}
Therefore, we have
\begin{gather}\label{eq:grad_F_pi_first}
\begin{split}
2\nabla_{ij} F_{M}(\Y) &= -\langle \what,\, (\Dij\Ky)\,\what\rangle +\\
&\qquad -\,2\left(\,\langle\what,\, \Dij\bm{v}_0(\Y)\rangle - \langle\what,\,(\Dij\Ky)\,\what\rangle\,\right)\\
&= -2\langle \what,\,\Dij \bm{v}_0(\Y)\rangle + \langle\what,\,(\Dij\Ky)\,\what\rangle.
\end{split}
\end{gather}

Now recall that
\begin{equation*}
    [\Dij\Ky]_{m_1,m_2} = \Dij\kappa(y_{m_1}, y_{m_2}),
\end{equation*}
for all $m_1,\,m_2 \in [M]$. Therefore, under \Cref{assumption:gradient_kappa},
\begin{equation}
\nabla_{ij} \kappa(y_{m_1},y_{m_2}) = \delta_{i,m_1}(y_{m_2j} - \yij)\bar{\kappa}(\yi,y_{m_2}) + \delta_{i,m_2}(y_{m_1j} - \yij)\bar{\kappa}(y_{m_1}, \yi).
\end{equation}

First, we denote $[\bm{a}\otimes\bm{b}]_{ij} = a_ib_j$ as the outer product between vectors $\bm{a}$ and $\bm{b}$ with possibly different dimensions. We also denote $\bm{a}\odot\bm{b}$ as the elementwise (i.e., Hadamard) product between $\bm{a}$ and $\bm{b}$ with identical dimensions. Then, we have
\begin{equation*}
        \nabla_{ij} \Ky = \bij\otimes \bm{e}_i + \bm{e}_i\otimes \bij,
\end{equation*}
where $\bm{e}_i$ is the $i$th elementary vector, i.e., $[\bm{e}_i]_{m} = \delta_{im},$ and we define
\begin{equation*}
        \bij := (\Y_j - \yij\bm{1})\odot \Ki = \Y_{j}\odot \Ki - \yij \Ki,
\end{equation*}
with $\Y_{j} \in \RR^{M}$ is the vector containing the $j$-th entry of each $\yi$, identical to the $j$th column of the matrix $\Y$, and $\Ki = \Kmbar(\Y)\bm{e}_i$. Then,
\begin{align}\label{eq:w_hat_grad_K_w_hat_final}
\langle\what,\,(\Dij\Ky)\,\what\rangle &= \langle\what,\, (\bij\otimes\bm{e}_i + \bm{e}_i\otimes\bij)\what\rangle\nonumber\\
&= 2 \whati \langle \what,\,\bij\rangle \nonumber \\
&= 2 \whati \Big( \langle\what,\,\Y_{j}\odot \Ki - \yij \Ki\rangle\Big) \nonumber \\
&= 2 \whati \Big( \langle\what,\, \Y_{j}\odot \Ki\rangle - \yij\langle\what,\, \Ki\rangle \Big)
\end{align}
 For $m \in [M]$, we recall that $\nabla\kappa$ is bounded on $\X\times\X$ to prove
 \begin{align*}
\Dij \vzerov(\Y) &= \left(\Dij \vzero(\yi)\right)\bm{e}_i\\
\Dij \vzero(\yi)&= \int_{\X} \Dij \kappa( x,\yi) \dpix \\
& = \int_{\X}  (\xj-\yij)\bar{\kappa}(x,\yi) \dpix\\
& = \int_{\X}    \xj\bar{\kappa}( x,\yi) \dpix -\yij  \bvzero(\yi),
 \end{align*}
where $\xj\in\RR$ is the $j$th entry of integration variable $x\in\RR^d$, and we recall that $\bar{v}_{0}(y)= \int_{\X} \bar{\kappa}(x,y) \dpix$. We then have
\begin{equation*}
    \Dij \vzerov(\Y) =  \left( [\bvonev(\yi)]_{j}  - \yij\bvzero(\yi)\right)\bm{e}_i.
\end{equation*}

Thus,
\begin{equation}\label{eq:w_hat_grad_v_final}
    \langle\what,\,\nabla_{ij} \bm{v}_0(\Y)\rangle = \whati\left( [\bvonev(\yi)]_{j}  - \yij\bvzero(\yi)\right).
\end{equation}
Combining~\eqref{eq:grad_F_pi_first}, \eqref{eq:w_hat_grad_K_w_hat_final} and \eqref{eq:w_hat_grad_v_final}, we obtain 
\begin{align}\label{eq:2times_gradient_FM}
    2\Dij F_{M}(\Y) &= -2\whati\left([\bvonev(\yi)]_{j}  - \yij\bvzero(\yi)- \langle\what,\, \Y_{j}\odot\Ki\rangle + \yij\langle\what,\,\Ki\rangle
    \right) \nonumber \\
    &= -2\whati\left( [\bvonev(\yi)]_{j} - \yij \bvzero(\yi) + \yij \langle\what,\,\Ki\rangle - \langle\what,\,\Y_{j}\odot\Ki\rangle 
    \right)
\end{align}
Now, 
\begin{align}\label{eq:ubar_i}
   - \yij \bvzero(\yi) + \yij \langle\what,\,\Ki\rangle &= \yij \left( \sum\limits_{m=1}^{M}\whatms\bar{\kappa}(\yi,\yms) - \int_{\X} \bar{\kappa}(x,\yi) \dpix \right)\nonumber\\
   &= (\hvonem(\Y))_{i,j} - [\bvonev(\yi)]_j
\end{align}
matching the definition of $\hvonem$ in \cref{eq:v_hat_def}. Using the symmetry of $\bar{\kappa}$.%
\begin{align}\label{eq:what_yj_hadamard_K}
    \langle\what,\, \Y_{j}\odot \Ki\rangle &= \sum_{m=1}^M \whatms\ymsj\bar{\kappa}(\yms,\yi) \nonumber\\
    &= \sum_{m=1}^M \bar{\kappa}(\yi,\yms)\whatms\ymsj \nonumber\\
    &= \langle\bm{e}_i,\, \Kmbar(\Y)\bm{W}(\Y)\Y_j\rangle \nonumber\\
    &= \left[\,\Kmbar(\Y)\bm{W}(\Y)\Y\right]_{i,j}
\end{align}
where $\bm{W}(\Y)\in\mathbb{R}^{M\times M}$ satisfies $[\bm{W}(\Y)]_{i,m} = \whati\delta_{i,m}$.  Combining \eqref{eq:ubar_i}, \eqref{eq:what_yj_hadamard_K}, and the fact that $[\bvonev(\yi)]_j$ is the $(i,j)$-th entry of the matrix $\bvonem(\Y)$, we conclude that
\begin{equation*}
    \left([\bvonev(\yi)]_{j} - \yij\bvzero(\yi) + \yij \langle\what,\, \Ki\rangle - \langle\what,\, \Y_{j}\odot \Ki\rangle 
    \right)
\end{equation*}
is the $(i,j)$-entry of the matrix $\hvonem(\Y) -\bm{\bar{K}}(\Y) \bmW \Y$. Thus, \eqref{eq:2times_gradient_FM} is
the $(i,j)$-entry of the matrix 
\begin{equation}
    2 \bmW \left( \bm{\bar{K}}(\Y) \bmW \Y  - \hvonem(\Y)\right).
\end{equation}
In other words,
\begin{align}
    \nabla F_{M}(\Y) &= \bmW \left( \bm{\bar{K}}(\Y) \bmW \Y  - \hvonem(\Y)\right).
\end{align}%
When the matrices $\bmW$ and $\bm{\bar{K}}(\Y)$ are nonsingular, we have
\begin{equation}
    \nabla F_{M}(\Y) = \bmW \bm{\bar{K}}(\Y) \bmW\left( \Y - \bmW^{-1} \bm{\bar{K}}(\Y)^{-1} \hvonem(\Y) \right).
\end{equation}
Finally, observe that the function $\bm{\Psi}_{\MMS}$ defined by \eqref{eq:fixed_point_general_case} satisfies
\begin{equation}
\bm{\Psi}_{\MMS}(\Y) = \bmW^{-1} \bm{\bar{K}}(\Y)^{-1} \hvonem(\Y).
\end{equation}

\subsection{Proof of \texorpdfstring{\Cref{prop:mmd_inimization_using_msip}}{Proposition \ref{prop:mmd_inimization_using_msip}}}\label{proof:mmd_inimization_using_msip}
\newcommand{\Rsd}{\mathbb{R}_{+}^{d}}

Let $\Rsd = \{x\in\RR^d\,|\,x_j > 0\,\forall j\in[M]\}$.

\begin{lemma}\label{lemma:open_inv_kernel}
The set $A = \{Y\in\RR^{M\times d}\,|\,\det \Ky| > 0\}$ is open and nonempty for the squared exponential kernel given any finite bandwidth $s> 0$.
\end{lemma}
\begin{proof}
    Since $\Y \mapsto \bm{K}(\Y)$ and the determinant are both continuous functions, their composition $Y\mapsto \det \Ky$ is continuous as well. Thus, $A$ is open as the pre-image of an open set under a continuous function. Now set $r := s \log M$ and choose $Y$ such that $\|y_i-y_j\|>r$ for $i\neq j$ (e.g., $y_j = j (1+r) \bm{e}_1$ for elementary vector $\bm{e}_1$). We know then that, for $i\neq j$,
    \[\kappa(y_i,y_j) = \exp(-\|y_i-y_j\|^2/s^2) < \exp(-r^2/s^2) \leq \frac{1}{M}.\]
    Moreover, we have $\kappa(y_i,y_i) = 1$ for $i \in [M]$. Then, we use Gershgorin's circle theorem by noting the radius of every Gershgorin circle is bounded by $(M-1)/M < 1$. More precisely, this proves that the absolute value of any eigenvalue of $\Ky$ is bounded from below by $1/M$, and thus ensuring $\det \Ky > 0$. Therefore, we know that $Y\in A$ and thus $A\neq\emptyset$.
\end{proof}

\begin{proof}[Proof of \Cref{prop:mmd_inimization_using_msip}]
    We recall that $\kappa$ is the squared-exponential kernel and proceed with a simple topological argument. For any $x\in\Rsd$, we define %
    $r_x := \min_j x_j$  to describe a ball $B_{r_x}\subset\Rsd$ centered on $x$, so we prove that $\Rsd$ is an open set. %
    Further, we use $A$ as given in \cref{lemma:open_inv_kernel} to define the open set of configurations $Y$ with invertible kernel matrices. %
    Finally, we know that $\vzerov\in\mathcal{H}$ and, thus, it must be continuous.

    Now, recall that $\hat{\bm{w}}:A\to\RR^M$ is defined as $\hat{\bm{w}}(\Y) = \Ky^{-1}\vzerov$, and observe that $\hat{\bm{w}}$ is continuous. Indeed, $\Y \mapsto \Ky^{-1}$ is continuous by the continuity of $\Y \mapsto \Ky$ and the continuity of the inverse; and $\Y \mapsto \vzerov(\Y)$ is continuous on $A$, since $\vzerov \in \mathcal{H}$ and thus continuous. In the following, define $B := \{Y\in A\,|\, \hat{\bm{w}}(Y) \in\Rsd\}$. Since $\Rsd$ is open and $\hat{\bm{w}}$ is continuous, $B$ is open as pre-image of an open set by a continuous function.

    Consider the MSIP setting with some $\Y^{(t)}\in A$ such that $\bm{w}(\yt)\in\Rsd$ and let $U^{(t)} = \bm{P}(\yt)^{-1}\nabla F_M(\yt)=\bm{\Psi}_{\MMS}(Y^{(t)})$. %
    We know that $\bm{P}(\yt)=\bmW\Ky\bmW$ is positive definite, as $\bm{x}^\top \bm{P}(\yt)\bm{x} = (\bmW \bm{x})^\top \Ky (\bmW\bm{x}) > 0$ for every vector $\bm{x}\neq 0$. This comes from the fact that $\bmW$ is invertible and $\Y\in A$, so $\Ky$ is positive definite, giving that both $\bm{P}(\yt)$ and $\bm{P}(\yt)^{-1}$ are positive definite. Moreover, $-\bm{P}(\yt)^{-1} \nabla F_M(\yt)$ is a descent direction of the function $F_M$ since %
    \begin{align*}
        \langle \nabla F_M(\yt), -\bm{P}(\yt)^{-1}&\nabla F_M(\yt)\rangle_F =\\ &\sum_{i=1}^M \langle(\nabla F_M(\yt))_{i,:}, -\bm{P}(\yt)^{-1}(\nabla F_M(\yt))_{i,:}\rangle < 0
    \end{align*}
    for the Frobenius inner product $\langle\cdot,\cdot\rangle_F$. %
    Thus, there exists some $\eta_1$ such that $F_M(\yt - \eta_1 U^{(t)}) < F_M(\yt)$. As $B$ is an open set, there exists $\eta_2>0$ such that $\yt-\eta_2 U^{(t)}\in A$ and nonzero $\eta_3 < \eta_2$ such that $\bm{w}(\yt - \eta_3 U^{(t)})\in \Rsd$. Step sizes $\eta_2$ and $\eta_3$ come from the openness of $A$ and $B$, respectively. Thus, picking $\eta_{t} = \min(\eta_1,\eta_2,\eta_3)$, we get that $\ytplus = \yt - \eta_{t} U^{(t)}$ satisfies the proposition.
\end{proof}

\subsection{Proof of \texorpdfstring{\Cref{prop:exist_Y}}{Proposition \ref{prop:exist_Y}}}\label{app:proof_exists_config}
We first prove that there exists some bandwidth $s$ such that any configuration of unique points drawn from empirical $\pi$ satisfies $\hat{\bm{w}}(\y)\in\Rsd$. First, we admit the spectral decompostion of $\Ky$ as $U\Sigma U^\top$ for diagonal $\Sigma$ and orthogonal $U$. We know that
\[\Ky^{-1} = I - (I - \Ky^{-1}) = I - U(I - \Sigma^{-1})U^\top.\]
Therefore, we can see that
\[\Ky^{-1}\vzerov(\y) = \vzerov(\y) - U(I - \Sigma^{-1})U^\top\vzerov(\y).\]

What remains to be shown is that we can pick some bandwidth $s$ such that
\[\|U(I - \Sigma^{-1})U^\top\vzerov(\y)\|_{\infty} < \|\vzerov(\y)\|_\infty.\]
Using the topological equivalence bound $\|\cdot\|_\infty \leq \|\cdot\|_2$, we get that
\begin{align}
    \|U(I - \Sigma)U^\top\vzerov(\y)\|_{\infty} &\leq \|U(I - \Sigma^{-1})U^\top\vzerov(\y)\|_{2}\\
    &\leq \|(I - \Sigma^{-1})\|_2\|\vzerov(\y)\|_{2}\\
    &= \sqrt{M} (1-\sigma_{\mathrm{max}}^{-1}),
\end{align}
where $\sigma_{\mathrm{max}}$ is the maximum of the spectrum of $\Ky$, and the bound on $\|\vzerov(\y)\|_2$ comes from the fact that $\vzero(y) \leq 1$ (cf. \eqref{eq:positive_vzero}). Now, set $r := \min_{i\neq j} \|\yi - \yj\|^2_2$ and pick bandwidth
\[s = \frac{r}{\log (NM\sqrt{M})}.\]
Using Gershgorin's circle theorem, we know that, for any fixed $i\in[M]$,
\[\sigma_{\mathrm{max}} \leq \sum_{m=1}^M \kappa(\yi,\yms) \leq 1+(M-1)\exp(-s^{-1}r) = 1+\frac{M-1}{NM\sqrt{M}} < \frac{N\sqrt{M}+1}{N\sqrt{M}},\]
where we remark that this last inequality is \textit{strict}. Then,
\[\sqrt{M}(1-\sigma_{\mathrm{max}}^{-1}) < \sqrt{M}\left(1-\frac{N\sqrt{M}}{N\sqrt{M}+1}\right) = \frac{1}{N + M^{-1/2}} < \frac{1}{N} \leq \|\vzerov(\y)\|_\infty. \]
This last inequality comes from the fact that, for any $x_j$ drawn from the support of an empirical $\pi$, we have
\[\vzero(x_j) = \frac{1}{N}\sum_{i=1}^N \kappa(x_i,x_j) = \frac{1}{N} + \frac{1}{N}\sum_{i\neq j}\kappa(x_i,x_j) \geq \frac{1}{N}.\]
Since $\y$ were chosen from the support of $\pi$, we then know that $N^{-1} \leq \|\vzerov(\y)\|_\infty.$

\subsection{Proof of \texorpdfstring{\Cref{cor:ms_as_MSIP}}{Theorem \ref{cor:ms_as_MSIP}}}\label{proof:mean_shift_as_mmd_min}

We again define $C_\pi = \int_{\X} \int_{\X} \kappa(x,x^{\prime}) \dpix \mathrm{d}\pi(x^{\prime})$. First, for $w \in \mathbb{R}$ and $y \in \X$, we have
\begin{equation}
    \F(w\delta_{y}) = \frac{1}{2} \left( C_\pi- 2 w v_{0}(y) + w^2 \kappa(y,y) \right).
\end{equation}
Since this is a quadratic form on $w$, we get
\begin{equation}
    \hat{w}(y) = \argmin\limits_{w \in \mathbb{R}} \F(w\delta_{y}) = \frac{v_{0}(y)}{\kappa(y,y)}.
\end{equation}
Thus, by \eqref{eq:gradien_opt_mmd_under_assumption} in \Cref{thm:gradient_opt_mmd}, we get 
\begin{align}
    \nabla F_{1}(y) &=  \frac{\vzero(y)}{\kappa(y,y)}\bar{\kappa}(y,y)\frac{v_0(y)}{\kappa(y,y)}(y-\Psi_{\MMS}(y))\label{eq:def_Mone_MSIP_grad}\\
    &=\frac{v_{0}(y)}{\kappa(y,y)}\left(\bar{\kappa}(y,y) \frac{v_{0}(y)}{\kappa(y,y)} y - \hat{v}_1(y) \right).\label{eq:proof_cor_eq_1}
\end{align}
By definition \eqref{eq:v_hat_def}, we have
\begin{equation}\label{eq:app_v_hat_def}
    \hat{v}_1(y) = \left( \frac{v_{0}(y)}{\kappa(y,y)} \bar{\kappa}(y,y) - \bar{v}_{0}(y) \right) y + \bvonev(y).
\end{equation}
By rearranging \eqref{eq:app_v_hat_def}, we see
\begin{equation}\label{eq:proof_cor_eq_2}
  \frac{v_{0}(y)}{\kappa(y,y)} \bar{\kappa}(y,y) y - \hat{v}_1(y)  = \bar{v}_{0}(y) y - \bvonev(y).
\end{equation}
Combining \eqref{eq:proof_cor_eq_1} and \eqref{eq:proof_cor_eq_2}
\begin{align}
    \nabla F_{1}(y)  &=  \frac{v_{0}(y)}{\kappa(y,y)}\Big(\bar{v}_{0}(y) y - \bvonev(y) \Big).\\
    &= \frac{v_0(y)\bar{v}_0(y)}{\kappa(y,y)}(y-\Psi_{\mathrm{MS}}(y))\label{eq:def_Mone_MS_grad}
\end{align}

Now we prove that $\Psi_{\mathrm{MS}}$ and $\Psi_{\mathrm{MSIP}}$ are identical in the case that $\bar{\kappa} = \lambda\kappa$ for $\lambda\neq 0$. We first observe that
\begin{equation}
    \bar{v}_0(y) = \mathbb{E}_\pi[\bar\kappa(X,y)] = \mathbb{E}_{\pi}[\lambda\kappa(X,y)] = \lambda v_0(y).
\end{equation}
Then, we can see that
\begin{equation}
    \frac{v_0(y)\bar{v}_0(y)}{\kappa(y,y)} = \frac{v_0^2(y)}{\kappa^2(y,y)}\lambda\kappa(y,y) = \frac{v_0^2(y)}{\kappa^2(y,y)}\bar{\kappa}(y,y).
\end{equation}
Therefore, we can combine \cref{eq:def_Mone_MSIP_grad,eq:def_Mone_MS_grad} to arrive at the conclusion
\begin{equation}
    \Psi_{\mathrm{MS}}(y) = \Psi_{\mathrm{MSIP}}(y)
\end{equation}

\end{document}